\documentclass[10pt,twocolumn,letterpaper]{article}

\usepackage[pagenumbers]{cvpr} 

\usepackage[accsupp]{axessibility}  
\usepackage{graphicx}
\usepackage{amsmath}
\usepackage{amssymb}
\usepackage{booktabs}
\usepackage{makecell}
\usepackage[export]{adjustbox}
\usepackage[normalem]{ulem}
\usepackage{textpos}  %
\usepackage[table]{xcolor}

\usepackage{tabulary,multirow,overpic,xcolor,subfloat}

\usepackage[pagebackref,breaklinks,colorlinks, citecolor=citecolor, linkcolor=linkcolor]{hyperref}
\definecolor{citecolor}{HTML}{0071BC}
\definecolor{linkcolor}{HTML}{ED1C24}

\newcommand{\app}{\raise.17ex\hbox{$\scriptstyle\sim$}}

\newcolumntype{x}[1]{>{\centering\arraybackslash}p{#1pt}}
\newcolumntype{y}[1]{>{\raggedright\arraybackslash}p{#1pt}}

\newlength\savewidth\newcommand\shline{\noalign{\global\savewidth\arrayrulewidth
		\global\arrayrulewidth 1pt}\hline\noalign{\global\arrayrulewidth\savewidth}}
\newcommand{\tablestyle}[2]{\setlength{\tabcolsep}{#1}\renewcommand{\arraystretch}{#2}\centering\footnotesize}
\makeatletter\renewcommand\paragraph{\@startsection{paragraph}{4}{\z@}
	{.5em \@plus1ex \@minus.2ex}{-.5em}{\normalfont\normalsize\bfseries}}\makeatother

\DeclareMathAlphabet\mathbfcal{OMS}{cmsy}{b}{n}

\definecolor{Gray}{gray}{0.5}

\newcommand{\modelname}{FastInst\xspace}

\newcommand{\sota}[0]{state-of-the-art\xspace}

\definecolor{gain}{HTML}{34a853}
\newcommand{\gain}[1]{\textcolor{gain}{#1}}
\definecolor{lost}{HTML}{ea4335}

\newcommand{\figref}[1]{Figure~\ref{#1}}
\newcommand{\secref}[1]{Section~\ref{#1}}
\newcommand{\tabref}[1]{Table~\ref{#1}}
\newcommand{\appref}[1]{Appendix~\ref{#1}}

\newcommand{\apm}{AP}
\newcommand{\aps}{AP$_\text{S}$ & AP$_\text{M}$ & AP$_\text{L}$}
\newcommand{\apl}{AP$_\text{50}$ & AP$_\text{75}$ & AP$_\text{S}$ & AP$_\text{M}$ & AP$_\text{L}$}

\newcommand{\gr}{\rowcolor[gray]{.95}}
 
\usepackage[capitalize]{cleveref}
\crefname{section}{Sec.}{Secs.}
\Crefname{section}{Section}{Sections}
\Crefname{table}{Table}{Tables}
\crefname{table}{Tab.}{Tabs.}

\def\cvprPaperID{3555} 
\def\confName{CVPR}
\def\confYear{2023}

\begin{document}

\title{\modelname: A Simple Query-Based Model for Real-Time Instance Segmentation}
\author{Junjie He, Pengyu Li, Yifeng Geng, Xuansong Xie  \\
DAMO Academy, Alibaba Group\\
{\tt\small hejunjie.hjj@alibaba-inc.com, lipengyu007@gmail.com, cangyu.gyf@alibaba-inc.com} \\
{\tt\small xingtong.xxs@taobao.com}
}
\maketitle

\begin{abstract}
Recent attention in instance segmentation has focused on query-based models. Despite being non-maximum suppression (NMS)-free and end-to-end, the superiority of these models on high-accuracy real-time benchmarks has not been well demonstrated. In this paper, we show the strong potential of query-based models on efficient instance segmentation algorithm designs. We present FastInst, a simple, effective query-based framework for real-time instance segmentation. FastInst can execute at a real-time speed (\ie, 32.5 FPS) while yielding an AP of more than 40 (\ie, 40.5 AP) on COCO \texttt{test-dev} without bells and whistles.
Specifically, FastInst follows the meta-architecture of recently introduced Mask2Former. Its key designs include instance activation-guided queries, dual-path update strategy, and ground truth mask-guided learning, which enable us to use lighter pixel decoders, fewer Transformer decoder layers, while achieving better performance. The experiments show that FastInst outperforms most state-of-the-art real-time counterparts, including strong fully convolutional baselines,  in both speed and accuracy. Code can be found at \url{https://github.com/junjiehe96/FastInst}.
\end{abstract}

\section{Introduction}

Instance segmentation aims to segment all objects of interest in an image. The mainstream methods like Mask R-CNN~\cite{he2017mask,huang2019mask,chen2019hybrid,liu2018path} follow the design of detection-then-segmentation. Despite being simple and intuitive, those methods generate a lot of duplicate region proposals that introduce redundant computations. To improve efficiency, many single-stage methods~\cite{yolact-plus-tpami2020,lee2020centermask,chen2019tensormask,wang2020solov2} built upon Fully Convolutional Networks (FCNs)~\cite{long2015fully} appear. They segment objects end-to-end without region proposals. The inference speed of such methods is appealing, especially in real-time scenes. 
However, due to the dense predictions, the classical single-stage methods still rely on manually-designed post-processing steps like non-maximum suppression (NMS).

\begin{figure}[t]
	\centering
	\includegraphics[width=\linewidth]{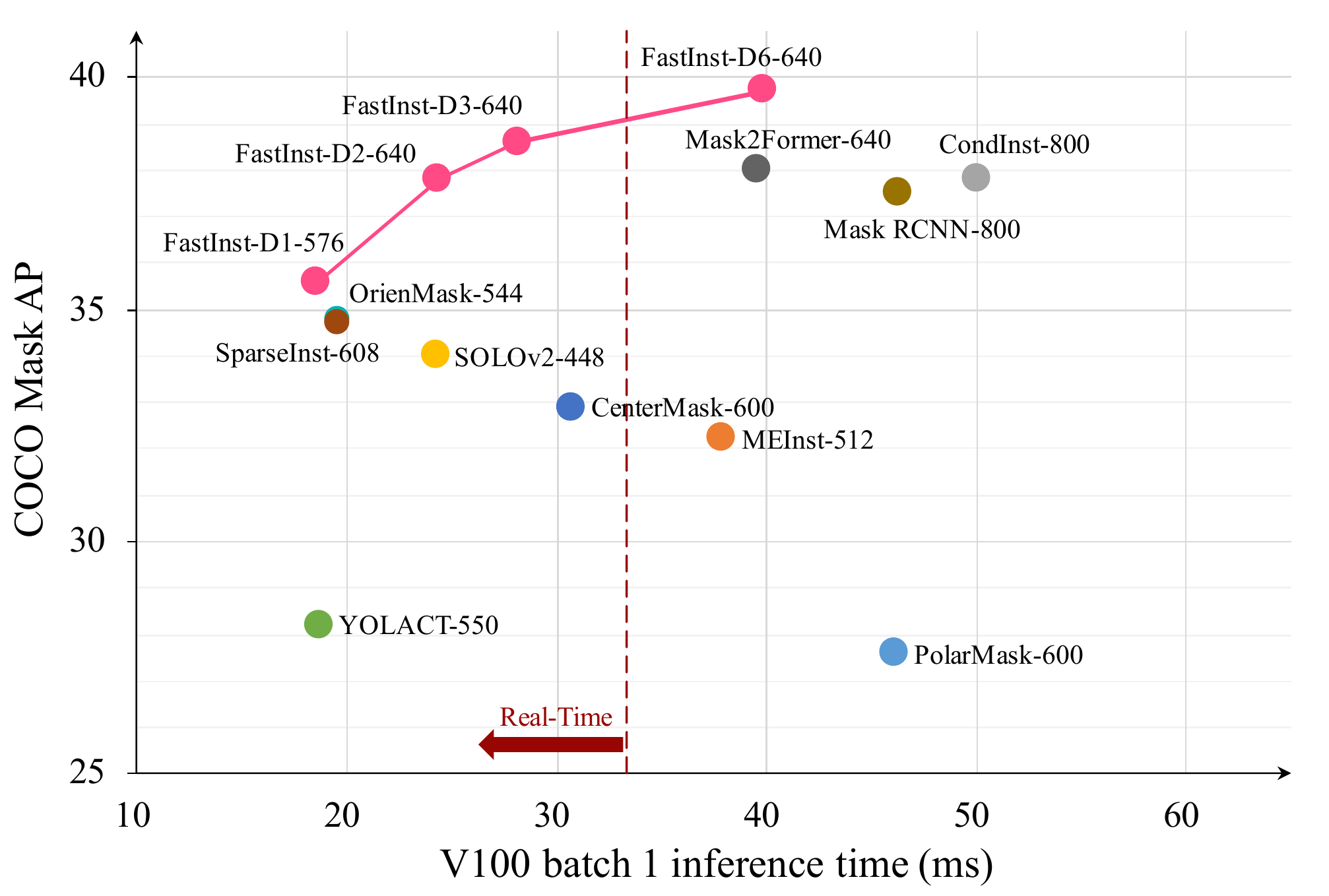}
	\caption{\textbf{Speed-performance trade-off on COCO \texttt{test-dev}.} All models employ ResNet-50~\cite{he2016deep} as the backbone except OrienMask with DarkNet-53~\cite{redmon2018yolov3}. Our \modelname surpasses most \sota real-time instance segmentation algorithms in both speed and accuracy. To keep the speed and accuracy in a similar order, Mask2Former here takes the pyramid pooling module-based~\cite{zhao2017pspnet} FPN as the pixel decoder, the same as FastInst and SparseInst. }
	\label{fig:tradeoff}
	\vspace{-5mm}
\end{figure}

Recently, with the success of DETR~\cite{detr} in object detection, query-based single-stage instance segmentation methods~\cite{li2021panopticsegformer,cheng2021maskformer,cheng2021mask2former, wu2022efficient} have emerged. Instead of convolution, they exploit the versatile and powerful attention mechanism~\cite{vaswani2017attention} combined with a sequence of learnable queries to infer the object class and segmentation mask. For example, Mask2Former~\cite{cheng2021mask2former} simplifies the workflow of instance segmentation by adding a pixel decoder and a masked-attention Transformer decoder on top of a backbone. Unlike previous methods~\cite{he2017mask,wang2020solov2}, Mask2Former does not require additional handcrafted components, such as training target assignment and  NMS post-processing. While being simple, Mask2Former has its own issues: 
(1) it requires a large number of decoder layers to decode the object queries since its queries are learned static and need a lengthy process to refine; 
(2) It relies upon a heavy pixel decoder, \eg, multi-scale deformable attention Transformer (MSDeformAttn)~\cite{zhu2021deformable}, because its object segmentation mask straightforwardly depends on the output of the pixel decoder, which is used as a per-pixel embedding feature for distinguishing different objects;
(3) masked attention restricts the receptive field of each query, which may cause the Transformer decoder to fall into a suboptimal query update process. 
Although Mask2Former achieves outstanding performance, its superiority on fast, efficient instance segmentation has not been well demonstrated, which yet is critical for many real-world applications such as self-driving cars and robotics.
In fact, due to the lack of prior knowledge and the high computational complexity of the attention mechanism, the efficiency of query-based models is generally unsatisfactory ~\cite{cheng2021mask2former, li2021panopticsegformer,hu2021istr}. 
The efficient real-time instance segmentation benchmarks are still dominated by classical convolution-based models~\cite{cheng2022sparseInst,wang2020solov2}.

In this paper, we fill this gap by proposing \modelname, a concise and effective query-based framework for real-time instance segmentation. We demonstrate that the query-based model can achieve outstanding performance on the instance segmentation task while maintaining a fast speed, showing great potential in efficient instance segmentation algorithm design. 
As an example, our designed fastest query-based model with ResNet-50~\cite{he2016deep} backbone achieves 35.6 AP at 53.8 FPS (frames-per-second) on the COCO~\cite{lin2014coco} \texttt{test-dev}, evaluated on a single V100 GPU (see \figref{fig:tradeoff});
moreover, our best trade-off model can execute at a real-time speed, \ie, 32.5 FPS, while yielding an AP of more than 40, \ie, 40.5 AP, which to the best of our knowledge, has not yet been achieved in previous methods. 

Specifically, our model follows the meta-architecture of  Mask2Former~\cite{cheng2021mask2former}. To achieve efficient real-time instance segmentation, we have proposed three key techniques.
First, we use instance activation-guided queries, which dynamically pick the pixel embeddings with high semantics from the underlying feature map as the initial queries for the Transformer decoder. Compared with static zero~\cite{detr} or learnable~\cite{cheng2021maskformer,cheng2021mask2former} queries, these picked queries contain rich embedding information about potential objects and reduce the iteration update burden of the Transformer decoder. 
Second, we adopt a dual-path architecture in the Transformer decoder where the query features and the pixel features are updated alternately. Such a design enhances the representational ability of pixel features and saves us from the heavy pixel decoder design. Moreover, it makes a direct communication between query features and pixel features, which speeds up the iterative update convergence and effectively reduces the dependence on the number of decoder layers.
Third, to prevent the masked attention from falling into a suboptimal query update process, we introduce ground truth mask-guided learning. We replace the mask used in the standard masked attention with the last-layer bipartite matched ground truth mask to forward the Transformer decoder again and use a fixed matching assignment to supervise the outputs. This guidance allows each query to see the whole region of its target predicted object during training and helps masked attention attend within a more appropriate foreground region.

We evaluate \modelname on the challenging MS COCO dataset~\cite{lin2014coco}. As shown in \figref{fig:tradeoff}, \modelname obtains strong performance on the COCO benchmark while staying fast, surpassing most of the previous \sota methods. We hope \modelname can serve as a new baseline for real-time instance segmentation and advance the development of query-based instance segmentation models.

\section{Related Work}

Existing instance segmentation techniques can be grouped into three classes, \ie,  region-based methods, instance activation-based methods, and query-based methods.

\noindent\textbf{Region-based methods} first detect object bounding boxes and then apply RoI operations such as RoI-Pooling~\cite{Ren2015a} or RoI-Align~\cite{he2017mask} to extract region features for object classification and mask generation. As a pioneering work, Mask R-CNN~\cite{he2017mask} adds a mask branch on top of Faster R-CNN~\cite{Ren2015a} to predict the segmentation mask for each object. Follow-up methods either focus on improving the precision of detected bounding boxes~\cite{cai2018cascade,chen2019hybrid} or address the low-quality segmentation mask arising in Mask R-CNN~\cite{kirillov2020pointrend,tang2021look,chengwhl20}. Although the performance has been advanced on several benchmarks, these region-based methods suffer from a lot of duplicated region proposals that hurt the model's efficiency.

\noindent\textbf{Instance activation-based methods} employ some meaningful pixels to represent the object and train the features of these pixels to be activated for the segmentation during the prediction. A typical class of such methods is based on the center activation~\cite{zhang2022e2ec,wang2020solov2,yolact-plus-tpami2020,tian2020conditional}, which forces the center pixels of the object to correspond to the segmentation and classification. For example, SOLO~\cite{wang2020solo, wang2020solov2} exploits the center features of the object to predict a mask kernel for the segmentation. MEInst~\cite{zhang2020MEInst} and CondInst~\cite{tian2020conditional} build the model upon the center-activation-based detector FCOS~\cite{tian2021fcos} with an additional branch of predicting mask embedding vectors for dynamic convolution. Recently, SparseInst~\cite{cheng2022sparseInst} learns a weighted pixel combination to represent the object. The proposed \modelname exploits the pixels located in the object region with the high class semantics as the representation of the object and extracts their features as the queries.

\noindent\textbf{Query-based methods} have emerged with DETR~\cite{detr} and show that a convolutional backbone with an end-to-end set prediction-based Transformer encoder-decoder~\cite{vaswani2017attention} can achieve good performance on the instance segmentation task. SOLQ~\cite{dong2021solq} and ISTR~\cite{hu2021istr} exploit the learned object queries to infer mask embeddings for instance segmentation. 
Panoptic SegFormer~\cite{li2021panopticsegformer} adds a location decoder to provide object position information.
Mask2Former~\cite{cheng2021maskformer,cheng2021mask2former} introduces masked attention for improved performance and faster convergence.
Mask DINO~\cite{li2022mask} unifies object detection and image segmentation tasks, obtaining great results on instance segmentation.
Despite the outstanding performance, query-based models are usually too computationally expensive to be applied in the real world. Compared with convolutional networks~\cite{cheng2022sparseInst,wang2020solov2}, their advantages on fast, efficient instance segmentation have not been well demonstrated. Our goal is to leverage the powerful modeling capabilities of the Transformer while designing an efficient, concise, and real-time instance segmentation scheme to promote the application of query-based segmentation methods. In addition, many works~\cite{wang2021max,yu2022cmt} also utilize the dual-path Transformer architecture in image segmentation tasks. However, their designs are generally complex and hard to deploy. We build our dual-path architecture simply upon plain Transformer layers for improved efficiency.

\begin{figure*}[t]
	\centering
	\includegraphics[width=\linewidth]{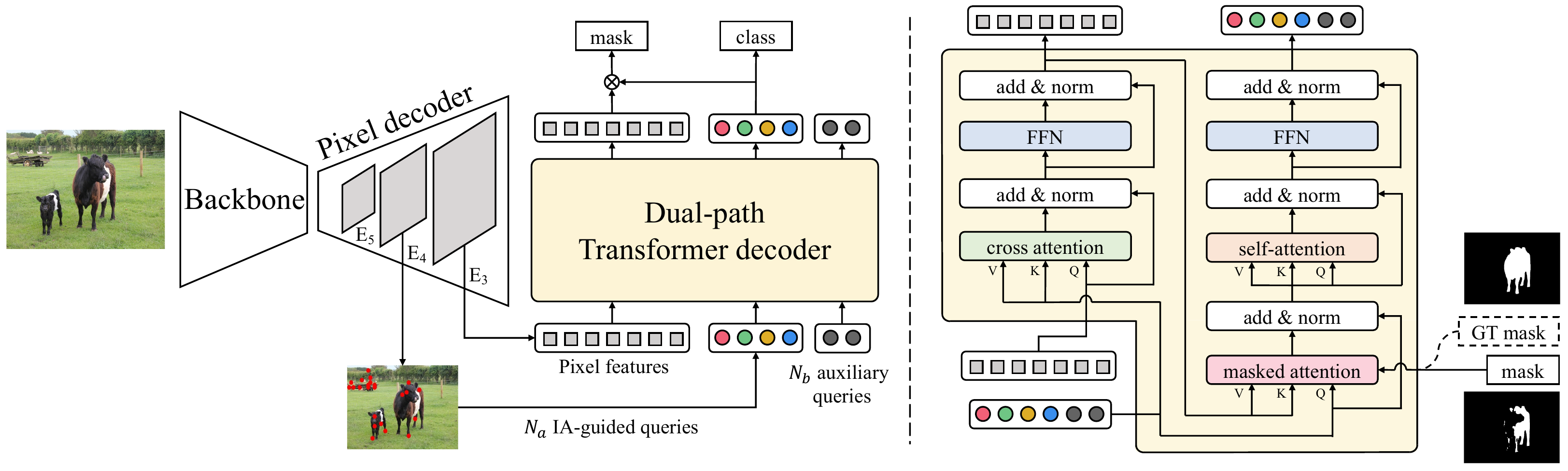}
	\caption{\textbf{Model overview.} FastInst consists of three modules: backbone, pixel decoder, and Transformer decoder. The backbone and pixel decoder extract and refine multi-scale features (\secref{sec:method:arch:pixel_decoder}). The Transformer decoder selects $N_a$ instance activation-guided queries (IA-guided queries) from the feature map E$_4$ (\secref{sec:method:arch:query}) and concatenates them with $N_b$ auxiliary learnable queries as initial queries. Then taking the initial queries and the flattened feature map E$_3$ as input, the Transformer decoder performs the object classification and segmentation at each layer with a dual-path update strategy (\secref{sec:method:arch:decoder}). During training, we introduce ground truth (GT) mask-guided learning to improve the performance of masked attention (\secref{sec:method:arch:gtmask}). For readability, we omit positional embeddings in this figure.
	}
	\label{fig:main_arch}
	\vspace{-10pt}
\end{figure*}

\section{Methods}

\subsection{Overall architecture}

As illustrated in \figref{fig:main_arch} , \modelname consists of three modules: backbone, pixel decoder, and Transformer decoder.

Our model feeds an input image $\mathbf{I}\!\in\!\mathbb{R}^{H \times W\times 3}$ to the backbone and obtains three feature maps C$_3$, C$_4$, and C$_5$, of which the resolutions are $1/8$, $1/16$, and $1/32$ of the input image, respectively. We project these three feature maps to the ones with 256 channels by a $1\!\times\!1$ convolutional layer and feed them into the pixel decoder. The pixel decoder aggregates the contextual information and outputs the enhanced multi-scale feature maps E$_3$, E$_4$, and E$_5$. After that, we pick $N_a$ instance activation-guided queries from the feature map E$_4$, concatenated with $N_b$ auxiliary learnable queries to obtain the total queries $\mathbf{Q}\in\!\mathbb{R}^{N\times256}$, where $N=N_a+N_b$. The  Transformer decoder takes as input the total queries $\mathbf{Q}$ as well as the flattened high-resolution pixel feature E$_3$, denoted as $\mathbf{X}\in\!\mathbb{R}^{L\times256}$, where $L=H/8\times W/8$. Then in the Transformer decoder, we update the pixel features $\mathbf{X}$ and the queries $\mathbf{Q}$ in a dual-path way and predict the object class and segmentation mask at each decoder layer. 

We now discuss each component in detail.

\subsection{Lightweight pixel decoder}
\label{sec:method:arch:pixel_decoder}

Multi-scale contextual feature maps are essential for image segmentation~\cite{kirillov2019panopticfpn,wang2020solov2,deeplabV2}. However, using a complicated multi-scale feature pyramid network increases the computational burden. Unlike previous methods~\cite{cheng2021maskformer,cheng2021mask2former}, which directly employ the underlying feature maps from the pixel decoder, we use the refined pixel features in the Transformer decoder to produce segmentation masks. 
This setup reduces the requirement of the pixel decoder for heavy context aggregation.
We thus can use a lightweight pixel decoder module. 
For a better trade-off between accuracy and speed, we use a variant called PPM-FPN~\cite{cheng2022sparseInst} instead of vanilla FPN~\cite{lin2016feature}, which adopts a  pyramid pooling module~\cite{zhao2017pspnet} after C$_5$ to enlarge receptive fields for improved performance. 

\subsection{Instance activation-guided queries}
\label{sec:method:arch:query}

Object queries play a crucial role in Transformer architecture~\cite{detr}.
One of the reasons for the slow convergence of DETR is that its object queries are zero-initialized. Although learnable queries~\cite{cheng2021mask2former} mitigate this issue, they are still image-independent and require many Transformer decoder layers to refine.
Inspired by Deformable DETR~\cite{zhu2021deformable}, which selects the query bounding boxes from pyramidal features for object detection, we propose instance activation-guided queries that straightforwardly pick the queries with high semantics from underlying multi-scale feature maps.
Specifically, given the output feature maps of the pixel decoder, we add an auxiliary classification head, followed by a softmax activation, on top of the feature map E$_4$ to yield the class probability prediction  $\mathbf{p}_i \!\in\!\Delta^{K \!+\!1}$ for each pixel, where $\Delta^{K+1}$ is the $(K+1)$-dimensional probability simplex, $K$ is the number of classes, added by one for ``no object" ($\varnothing$), $i$ is the pixel index, and the auxiliary classification head is composed of two convolutional layers with $3\!\times\!3$ and $1\!\times\!1$ kernel sizes, respectively.
Through $\mathbf{p}_i$ we obtain the foreground probability $p_{i,k_i}$, $k_i \!=\!\operatorname{argmax}_k\{p_{i,k}|p_{i,k}\!\in\!\mathbf{p}_i, k \!\in\!\{1,\cdots, K\}\}$ for each pixel. Then we select $N_a$ pixel embeddings from the feature map E$_4$ with high foreground probabilities as the object queries. Here we first select the ones with $p_{i,k_i}$ that is the \textit{local maximum} in the corresponding class plane (\ie, $p_{i,k_i}\!\geq\! p_{n,k_i}$, $n \!\in\!\delta(i)$, where $\delta(i)$ is the spatially 8-neighboring index set of $i$) and then pick the ones with the top foreground probabilities in $\{p_{i,k_i}\}_i$.
Note that a pixel with a non-local-maximum probability in the corresponding class plane means there exists a pixel in its 8-neighborhood which has a higher probability score of that class. With locations so close, we naturally prefer to pick its neighboring pixel rather than it as the object query. 

During training, we apply matching-based Hungarian loss~\cite{stewart2016,detr} to supervise the auxiliary classification head. 
Unlike~\cite{zhu2021deformable}, 
which employs prior anchor boxes and binary classification scores for the matching problem, 
we simply use the class predictions with a \textit{location cost} $\mathcal{L}_\text{loc}$ to compute the assignment costs. The location cost $\mathcal{L}_\text{loc}$  is defined as an indicator function that is 0 when the pixel is located in the region of that object; otherwise, it is 1. The intuition behind this cost is that only pixels that fall inside an object can have a reason to infer the class and mask embedding of that object. Also, the location cost reduces the bipartite matching space and speeds up training convergence. 

We term the queries generated from the above strategy as instance activation-guided (IA-guided) queries. Compared to the zero~\cite{detr} or learnable queries~\cite{cheng2021mask2former}, IA-guided queries hold rich information about potential objects at the initial and improve the efficiency of query iterations in the Transformer decoder. Note that we can also select the queries from feature maps E$_3$ or E$_5$. Larger feature maps contain richer instance clues but suffer heavier computational burdens. We use the middle-size feature map E$_4$ for a trade-off.

\subsection{Dual-path Transformer decoder}
\label{sec:method:arch:decoder}

After selecting $N_a$ IA-guided queries from the underlying feature map, we concatenate them with $N_b$ auxiliary learnable queries to obtain the total queries $\mathbf{Q}$, where auxiliary learnable queries are used to facilitate grouping background pixel features and provide general image-independent information in the subsequent dual update process. Then the total queries $\mathbf{Q}$ combined with the flattened $1/8$ high-resolution pixel features $\mathbf{X}$ are fed into the Transformer decoder.
In the Transformer decoder, we add positional embeddings for queries $\mathbf{Q}$ and pixel features $\mathbf{X}$,  followed by successive Transformer decoder layers to update them. One Transformer decoder layer \textit{contains} one pixel feature update and one query update.
The whole process is like an EM (Expectation–Maximization) clustering algorithm. E step: update pixel features according to the centers (queries) they belong to; M step: update cluster centers (queries).
Compared with the single-path update strategy~\cite{cheng2021mask2former}, the dual-path update strategy co-optimizes both pixel features and queries, reducing the dependence on heavy pixel decoders and acquiring more fine-grained feature embeddings.
Finally, we use the refined pixel features and queries to predict the object classes and segmentation masks at each layer.

\noindent\textbf{Positional embeddings.}
Location information is critical in distinguishing different instances with similar semantics, especially for objects with the same class~\cite{tian2020conditional,wang2020solo,wang2020solov2}. Instead of non-parametric sinusoidal positional embeddings~\cite{cheng2021mask2former}, we use the learnable positional embeddings, which we find can improve the model inference speed without compromising the performance. Specifically, we employ a fixed-size learnable spatial positional embedding $\mathbf{P}\in\!\mathbb{R}^{S\times S\times 256}$, where $S$ is the spatial size and we empirically set it to the rounded square root of the IA-guided query number $N_a$. During forwarding, we interpolate $\mathbf{P}$ to two different sizes. One is with the same size as E$_3$, which is then flattened as positional embeddings for pixel features $\mathbf{X}$; the other is with the same size as E$_4$, from which we select the positional embeddings for IA-guided queries according to their locations $\{(x_i,y_i)\}_{i=1}^{N_a}$ in the feature map E$_4$. The auxiliary learnable queries employ additional $N_b$ learnable positional embeddings.

\noindent\textbf{Pixel feature update.}
We first update the pixel features.
Given the flattened pixel features $\mathbf{X}$ and the queries $\mathbf{Q}$,  the pipeline of pixel feature update consists of a  cross-attention layer and a feedforward layer, as illustrated in the right side of \figref{fig:main_arch}.
The positional embeddings are added to queries and keys at every cross-attention layer~\cite{detr}.
For the update of pixel features, we do not use self-attention, which will introduce a massive computation and memory cost due to the long sequence length of pixel features. The global features can be aggregated through cross-attention on queries.

\noindent\textbf{Query update.}
Asymmetrically, we use masked attention followed by self-attention and feedforward network for the query update, as in Mask2Former~\cite{cheng2021mask2former}. 
Masked attention restricts the attention of each query to only attend within the foreground region of the predicted mask from the previous layer, and the context information is hypothesized to be gathered through following self-attention. Such a design has significantly improved query-based model performance in image segmentation tasks~\cite{cheng2021mask2former}.
Here the positional embeddings are also added to queries and keys at every masked- and self-attention layer. 

\noindent\textbf{Prediction.} We apply two separate 3-layer MLPs on top of refined IA-guided queries at each decoder layer to predict object classes and mask embeddings, respectively. Each IA-guided query needs to predict the probability of all object classes, including the "no object" ($\varnothing$) class. A linear projection is added to the refined pixel features to obtain mask features. Then mask embeddings are multiplied with mask features to obtain the segmentation masks for each query.
Here the parameters of MLPs and linear projection at each Transformer decoder layer are not shared, since queries and pixel features are updated alternately and their features can be in different representation spaces at different decoder layers. 
In addition, instance segmentation requires a confidence score for each prediction for evaluation. We follow previous work~\cite{cheng2021mask2former} and multiply the class probability score with the mask score  (\ie, the average of mask probabilities in the foreground region) as the confidence score.


\subsection{Ground truth mask-guided learning}
\label{sec:method:arch:gtmask}

Although masked attention introduces prior sparse attention knowledge that accelerates model convergence and improves the performance, it restricts the receptive field of each query and may cause the Transformer decoder to fall into a suboptimal query update process. To mitigate this issue, we introduce ground truth (GT) mask-guided learning. Firstly, we use the last layer's bipartite matched ground truth mask to replace the predicted mask used in $l$-th layer's masked attention. For the queries that do not match any instance in the last layer (including auxiliary learnable queries), we use the standard cross attention, \ie,
\begin{align}
	\mathbf{M}^l_i = \left\{\begin{array}{ll}
		\mathbf{M}^{gt}_j  & \text{if~} (i,j)\in\sigma \\
		\varnothing & \text{otherwise}
	\end{array}\right..
\end{align}
where $\mathbf{M}^l_i$ is the attention mask for the $i$-th query in the $l$-th layer,  $\sigma\!=\!\{(i,j)|i\in\{1, \cdots, N_a\}, j\in\{1, \cdots, N_{obj}\}\}$ is the matching of the last decoder layer, and $\mathbf{M}^{gt}_j$ is the matched ground truth mask for the $i$-th query in the last layer. Here $N_{obj}$ denotes the number of ground truth targets.
Then we use the replaced attention mask $\mathbf{M}^l$ combined with original output queries and pixel features of $l$-th layer, which are refined and for better guidance, as input to forward the  $l$-th Transformer decoder layer again. The new output is supervised according to the fixed matching $\sigma$, consistent with the last layer's bipartite matching results. This fixed matching ensures the consistency of the predictions at each Transformer decoder layer and saves the matching computation cost during training.
By such guided learning, we allow each query to see the whole region of its target predicted object during training, which helps the masked attention attend within a more appropriate foreground region.

\subsection{Loss function}
\label{sec:method:arch:loss}

The overall loss function for \modelname can be written as:
\begin{equation}
\mathcal{L} = \mathcal{L}_\text{IA-q}+\mathcal{L}_\text{pred}+\mathcal{L}'_\text{pred}
\end{equation}
where $\mathcal{L}_\text{IA-q}$ is the instance activation loss of the auxiliary classification head for IA-guided queries, $\mathcal{L}_\text{pred}$ and $\mathcal{L}'_\text{pred}$ are prediction loss and GT mask-guided loss, respectively.

\noindent\textbf{Instance activation loss.} The $\mathcal{L}_\text{IA-q}$ is defined as:
\begin{equation}
	\mathcal{L}_\text{IA-q} =\lambda_\text{cls-q}\mathcal{L}_\text{cls-q} 
\end{equation}
where $\lambda_\text{cls-q}$ is a hyperparameter and $\mathcal{L}_\text{cls-q}$ is the  cross-entropy loss. We use the Hungarian algorithm~\cite{kuhn1955hungarian} to search for the optimal bipartite matching between the prediction and ground truth sets. For the matching cost, we add an additional location cost $\mathcal{L}_\text{loc}$ of a weight $\lambda_\text{loc}$ to the above classification cost, as illustrated in \secref{sec:method:arch:query}.

\noindent\textbf{Prediction loss.} Following the prior work~\cite{cheng2021mask2former}, the prediction loss $\mathcal{L}_\text{pred}$ for the Transformer decoder is defined as:
\begin{equation}
	\mathcal{L}_\text{pred} = \sum_{i=0}^{D}  (\lambda_\text{ce}\mathcal{L}^i_\text{ce} + \lambda_\text{dice} \mathcal{L}^i_\text{dice}) + \lambda_\text{cls} \mathcal{L}^i_\text{cls}
	\label{sec:loss:prediction}
\end{equation}
where $D$ denotes the number of Transformer decoder layers and $i \!=\!0$ represents the prediction loss for IA-guided queries before being fed into the Transformer decoder, $\mathcal{L}^i_\text{ce}$ and  $\mathcal{L}^i_\text{dice}$ denote the binary cross-entropy loss and dice loss~\cite{milletari2016v} for segmentation masks, respectively, and $\mathcal{L}_\text{cls}$ is the cross-entropy loss for object classification with a ``no object" weight of $0.1$. $\lambda_\text{ce}$, $\lambda_\text{dice}$, and $\lambda_\text{cls}$ are the hyperparameters to balance three losses. Similarly, we exploit the Hungarian algorithm to search for the best bipartite matching for target assignments. And for the matching cost, we add an additional location cost $\lambda_\text{loc} \mathcal{L}_\text{loc}$ for each query. 

\noindent\textbf{GT mask-guided loss.} The  GT mask-guided loss $\mathcal{L}'_\text{pred}$ is similar to Equation~\eqref{sec:loss:prediction}. The only differences are that it does not count the $0$-th layer's loss and uses a fixed target assignment strategy, which is consistent with the bipartite matching result of the last Transformer decoder layer.

\section{Experiments}

In this section, we evaluate \modelname on COCO~\cite{lin2014coco} and compare it with several state-of-the-art methods. We also conduct detailed ablation studies to verify the effectiveness of each proposed component.

\begin{table*}[t]
  \centering
  
  \tablestyle{4pt}{1.2}\scriptsize\begin{tabular}{y{58} | x{42} | x{24} x{24} | x{30} | x{24}x{24}x{24} | x{24}x{24}x{24}}
  method & backbone & epochs & size & FPS & \apm & \apl \\
  \shline
  MEInst~\cite{zhang2020MEInst} & R50 & 36 & 512 & 26.4 & 32.2 & 53.9 & 33.0 & 13.9 & 34.4 &48.7 \\
  CenterMask~\cite{lee2020centermask} & R50 & 48	& 600 & 32.6 & 32.9	& - & -	& 12.9	& 34.7 & 48.7 \\
  SOLOv2~\cite{wang2020solov2} & R50 & 36 & 448 & 41.3 & 34.0 & 54.0 & 36.1 & 10.3 & 36.3 & 54.4 \\
  OrienMask~\cite{du2021realtime} & D53 & 100 & 544 & 51.0 & 34.8 & 56.7 & 36.4 & 16.0 & 38.2 & 47.8  \\
  SparseInst~\cite{cheng2022sparseInst} & R50 & 144 & 608 & 51.2 & 34.7 & 55.3 & 36.6 & 14.3 & 36.2 & 50.7 \\	
  YOLACT~\cite{yolact-iccv2019} & R50 & 54 & 550 & 53.5 & 28.2 & 46.6 & 29.2 & \phantom{0}9.2 & 29.3 & 44.8 \\
  \gr
  \textbf{\modelname-D1} (ours) & R50 & 50 & 576 & \textbf{53.8} & 35.6 & 56.7 & 37.2 & 8.8 & 53.0 & 72.8 \\
  CondInst~\cite{tian2020conditional} & R50 & 36 & 800 & 20.0	& 37.8	& 59.1	& 40.5 & \textbf{21.0}	& 40.3	& 48.7 \\
  Mask2Former$^{\text{\textdagger}}$ & R50 & 50 & 640 & 25.3 & 38.0 & \textbf{60.3} & 39.8 & 10.8 & 54.9 & 74.3 \\
  \gr
  \textbf{\modelname-D3} (ours) & R50 & 50 & 640 & 35.5 & \textbf{38.6} & 60.2 & \textbf{40.6} & 10.8 & \textbf{56.2} & \textbf{75.2} \\
  \hline
  YOLACT~\cite{yolact-iccv2019} & R101 & 54 & 700 & 33.5 & 31.2 & 50.6 & 32.8 & 12.1 & 33.3 & 47.1 \\
  \gr
  \textbf{\modelname-D1} (ours) & R101 & 50 & 640 & \textbf{35.3} & 38.3 & 60.2 & 40.5 & 10.7 & 55.8 & 74.8  \\
  SOLOv2~\cite{wang2020solov2} & R101 & 36 & 800 & 15.7 & 39.7 & 60.7 &  \textbf{42.9} & 17.3 &  42.9 & 57.4 \\
  CondInst~\cite{tian2020conditional} & R101 & 36 & 800 & 16.4 & 39.1 & 60.9 & 42.0 &  \textbf{21.5} & 41.7 & 50.9 \\
  Mask2Former$^{\text{\textdagger}}$ & R101 & 50 & 640 & 21.1 & 39.5 &  \textbf{61.7} & 41.6 & 11.2 & 56.3 & 75.8 \\
  \gr
  \textbf{\modelname-D3} (ours) & R101 & 50 & 640 & 28.0 &  \textbf{39.9} &  61.5 & 42.3 & 11.4 &  \textbf{57.1} &  \textbf{76.6} \\
  \hline
  SOLOv2~\cite{wang2020solov2} & R50-DCN & 36 & 512 & 32.0 & 37.1 & 57.7 & 39.7 & 12.9 & 40.0 & 57.4 \\
  YOLACT++~\cite{yolact-plus-tpami2020} & R50-DCN & 54 & 550 & 39.4 & 34.1 & 53.3 & 36.2 & 11.7 & 36.1 & 53.6 \\
  SparseInst~\cite{cheng2022sparseInst} & R50-d-DCN & 144 & 608 & 46.5 & 37.9 & 59.2 & 40.2 & \textbf{15.7} & 39.4 & 56.9 \\
  \gr
  \textbf{\modelname-D1} (ours) & R50-d-DCN & 50 & 576 & \textbf{47.8} & 38.0 & 59.7 & 39.9 & 10.0 & 54.9 & 74.5  \\
  \gr
  \textbf{\modelname-D3} (ours) & R50-d-DCN & 50 & 640 & 32.5 & \textbf{40.5} & \textbf{62.6} & \textbf{42.9} & 11.9 & \textbf{57.9} & \textbf{76.7} \\
  
  \end{tabular}

   \caption{\textbf{Instance segmentation on COCO \texttt{test-dev}.} \modelname outperforms most previous real-time instance segmentation algorithms in both accuracy and speed.
   	 Mask2Former$^{\text{\textdagger}}$ denotes a light version of Mask2Former~\cite{cheng2021mask2former} that exploits PPM-FPN as the pixel decoder, as in \modelname and SparseInst~\cite{cheng2022sparseInst}.  \modelname-D$\alpha$ represents \modelname with $\alpha$ Transformer decoder layers. ``R50-d-DCN" means ResNet-50-d~\cite{he2019bag} backbone with deformable convolutions~\cite{zhu2019deformable}.
   	 For a fair comparison, all entries are \textit{single-scale} results.}


\label{tab:insseg:coco}
\end{table*}

\subsection{Implementation details}
\label{sec:exp:impl}

Our model is implemented using Detectron2~\cite{wu2019detectron2}. We use the AdamW~\cite{loshchilov2018decoupled} optimizer with a step learning rate schedule. The initial learning rate is 0.0001, and the weight decay is 0.05. We apply a learning rate multiplier of 0.1 to the backbone, which is ImageNet-pretrained, and decay the learning rate by 10 at fractions 0.9 and 0.95 of the total number of training iterations. 
Following \cite{cheng2021mask2former}, we train our model for 50 epochs with a batch size of 16. For data augmentation, we use the same scale jittering and random cropping as in \cite{cheng2022sparseInst}. For example, the shorter edge varies from 416 to 640 pixels, and the longer edge is no more than 864 pixels. We set the loss weights $\lambda_\text{cls}$, $\lambda_\text{ce}$, and $\lambda_\text{dice}$ to 2.0, 5.0, and 5.0, respectively, as in \cite{cheng2021mask2former}. $\lambda_\text{cls-q}$ and $\lambda_\text{loc}$ are set to 20.0 and 1000.0, respectively. We use 100 IA-guided queries and 8 auxiliary learnable queries by default. We report the AP performance as well as the FLOPs and FPS. FLOPs are averaged using 100 validation images. FPS is measured on a V100 GPU with a batch size of 1 using the entire validation set. Unless specified, we use a shorter edge of 640 pixels with a longer edge not exceeding 864 pixels to test and benchmark models.

\subsection{Main results}
We compare \modelname with state-of-the-art methods on the COCO dataset in \tabref{tab:insseg:coco}.
Since the goal of \modelname is for an efficient real-time instance segmenter, we mainly compare it with state-of-the-art real-time instance segmentation algorithms in terms of accuracy and inference speed. The evaluation is conducted on COCO \texttt{test-dev}. We provide \modelname with different backbones and different numbers of Transformer decoder layers to achieve a trade-off between speed and accuracy. The results show that \modelname outperforms most previous \sota real-time instance segmentation methods with better performance and faster speed. For example, with a ResNet-50~\cite{he2016deep} backbone, the designed \modelname-D1 model outperforms a strong convolutional baseline SparseInst~\cite{cheng2022sparseInst} by 0.9 AP while using fewer training epochs and less inference time. We also compare \modelname with the query-based model Mask2Former~\cite{cheng2021mask2former}. To keep the speed and accuracy in a similar order, we replace the MSDeformAttn~\cite{zhu2021deformable} pixel decoder in Mask2Former with the PPM-FPN-based one, which is the same as \modelname as well as SparseInst~\cite{cheng2022sparseInst}. Meanwhile, for a fair comparison, the training setting of Mask2Former, including data augmentation, is replaced with the same as \modelname. As expected, Mask2Former relies on a strong pixel decoder and performs worse than \modelname in both accuracy and speed with a lighter pixel decoder (even if it has 9 decoder layers), showing less efficiency in the real-time benchmark. Besides, with ResNet-50-d-DCN~\cite{he2019bag,zhu2019deformable} backbone, our algorithm achieves 32.5 FPS and 40.5 AP, the only algorithm in \tabref{tab:insseg:coco} with AP $\!>\!40$ while maintaining a real-time speed ($\geq\!$ 30 FPS). \figref{fig:tradeoff} illustrates the speed-accuracy trade-off curve, which also demonstrates the superiority of our method.

\begin{table}[t]
	\centering
	\tablestyle{2pt}{1.2}
	\scriptsize
	\begin{tabular}{p{40px} | x{18} | x{25} x{21} x{21}x{21}|x{26}x{21}}
		 & $D$ & \phantom{00}\apm$^\texttt{val}$ & \aps & \phantom{0}FLOPs & FPS \\
		\shline
		 zero~\cite{detr} & 1 & \phantom{0}31.5 & 10.8 & 33.6 & 52.6  & \phantom{0}58.4G & 50.0 \\
		\hline
		learnable~\cite{cheng2021mask2former} & 1 &  \phantom{0}34.6  & 13.5 & 37.5 & 55.4 & \phantom{0}58.4G & 50.0 \\		
		resize~\cite{pei2022osformer} & 1 & \phantom{0}34.9  & 13.7 & 37.9 & 56.2 & \phantom{0}58.4G & 49.7 \\
		\hline
		\textbf{IA-guided} & 1 & \phantom{0}\textbf{35.6}  & \textbf{14.3} & \textbf{38.8} & \textbf{56.6} & \phantom{0}59.6G & 48.8 \\ 
		\hline\hline
		 zero~\cite{detr} & 3 & \phantom{0}37.2 & 15.4 & 40.3 & 58.6 & \phantom{0}74.3G & 36.1 \\
		 \hline
		learnable~\cite{cheng2021mask2former} & 3 & \phantom{0}37.5 & 15.0 & 40.6 & 59.0 & \phantom{0}74.3G & 36.1 \\
		resize~\cite{pei2022osformer} & 3 & \phantom{0}37.6  & 15.6 & 40.4 & 59.7 & \phantom{0}74.3G & 36.0 \\
		\hline
		\textbf{IA-guided} &  3 & \phantom{0}\textbf{37.9}  & \textbf{16.0} & \textbf{40.7} & \textbf{60.1} & \phantom{0}75.5G & 35.5 \\
		\hline
	\end{tabular}
\caption{\textbf{IA-guided queries.} Our IA-guided queries perform better than other methods, especially when the Transformer decoder layer number (\ie, $D$) is small.}
\vspace{-3mm}
\label{tab:ablation:iaquery}
\end{table}

\begin{table}[t]
	\centering
	\tablestyle{2pt}{1.2}
	\scriptsize
	\begin{tabular}{p{75px} | x{10} | x{21} x{17} x{17}x{17}|x{24}x{20}}
		& $D$ & \phantom{00}\apm$^\texttt{val}$ & \aps & \phantom{0}FLOPs & FPS \\
		\shline
		single pixel feat. update & 6 & \phantom{0}32.5  & 13.9 & 36.3 & 50.5 & \phantom{0}85.4G & 35.5  \\
		single query update~\cite{cheng2021mask2former} & 6 & \phantom{0}36.9  & 15.0 & 39.6 & 59.8 & \phantom{0}63.3G & 35.0 \\
		\hline
		dual query-then-pixel & 3 & \phantom{0}37.8 & \textbf{16.0} & 40.6 & 60.0 & \phantom{0}75.5G & 35.5  \\
		\hline
		\textbf{dual pixel-then-query} & 3 & \phantom{0}\textbf{37.9}  & \textbf{16.0} & \textbf{40.7} & \textbf{60.1} & \phantom{0}75.5G & 35.5  \\
		\hline
	\end{tabular}
	\caption{\textbf{Dual-path update strategy.} Our dual pixel-then-query update strategy consistently outperforms single-path update strategies. We double the Transformer decoder layers (\ie, $D$) for the single-path update strategies for a fair comparison.}
	\vspace{-3mm}
	\label{tab:ablation:dualpath}
\end{table}

\begin{table}[t]
	\centering
	\tablestyle{2pt}{1.2}
	\scriptsize
	\begin{tabular}{p{65px} | x{36} | x{20} x{15} x{15}x{15}|x{20}x{15}}
		& backbone & \phantom{0}\apm$^\texttt{val}$ & \aps & FLOPs & FPS \\
		\shline
		w/o GT mask guidance & R50 & 37.4 & 15.2 & 40.6 & 59.6 & 75.5G & 35.5 \\
		\textbf{w/  GT mask guidance} & R50 & \textbf{37.9} & \textbf{16.0} & \textbf{40.7} & \textbf{60.1} & 75.5G & 35.5 \\
		\hline\hline
		w/o GT mask guidance & R50-d-DCN & 39.7	& 17.5 & 43.1 & 61.9 & - & 32.5 \\
		\textbf{w/  GT mask guidance} & R50-d-DCN & \textbf{40.1} & \textbf{17.7} & \textbf{43.2} & \textbf{62.4} & - & 32.5 \\
		\hline
	\end{tabular}
	\caption{\textbf{GT mask-guided learning.} Our GT mask-guided learning improves the performance across different backbones.}
	\vspace{-3mm}
	\label{tab:ablation:gtmask}
\end{table}

\begin{table}[t]
	\centering
	\tablestyle{2pt}{1.2}
	\scriptsize
	\begin{tabular}{y{72} | x{25} x{20} x{20}x{20}|x{24}x{20}}
		& \phantom{0}\apm$^\texttt{val}$ & \aps & FLOPs & FPS \\
		\shline
		FPN~\cite{lin2016feature} & 37.4 & 15.5 & 40.3 & 59.4 & 75.4G & 35.7 \\
		\hline
		Transformer-Encoder~\cite{detr} & 38.9 & 17.3 & 41.6 & 61.4 & 78.5G & 29.5 \\
		MSDeformAttn~\cite{zhu2021deformable} & \textbf{40.0} & \textbf{17.5} & \textbf{43.5} & \textbf{62.2} & 114.7G & 21.2 \\				
		\hline
		\textbf{PPM-FPN}~\cite{cheng2022sparseInst} & 37.9  & 16.0 & 40.7 & 60.1 & 75.5G & 35.5 \\
	\end{tabular}
	\caption{\textbf{Pixel decoder.} Stronger pixel decoders lead to higher performance but consume more computation. PPM-FPN obtains a better trade-off between accuracy and speed.}
	\vspace{-3mm}
	\label{tab:ablation:pixeldecoder}
\end{table}


\begin{table*}[t]
 	\begin{subtable}{0.32\linewidth}
  	\centering
  	\tablestyle{3pt}{1.2}
  	\scriptsize
  	\begin{tabular}{c | x{15}x{15}x{15}x{15} | x{20}x{15}}
  		$D$ & \apm$^\texttt{val}$ & \aps & FLOPs & FPS \\
  		\shline
		0 & 30.5 & 12.1 & 34.5 & 48.6 & 51.6G & 60.2 \\
		1 & 35.6 & 14.3 & 38.8 & 56.6 & 59.6G & 48.8 \\
		2 & 37.1 & 15.8 & 39.7 & 58.8 & 67.5G & 41.0 \\
		3 & 37.9 & 16.0 & 40.7 & 60.1 & 75.5G & 35.5 \\
		6 & 38.7 & \textbf{16.6} & 41.7 & 61.1 & 99.3G & 25.1 \\
		9 & \textbf{39.1} & 16.1 & \textbf{42.3} & \textbf{62.1} & 123.2G & 19.2 \\
  	\end{tabular}
  	\caption{\textbf{Transformer decoder layer number.} \modelname benefits from more Transformer decoder layers. }
  	\label{tab:ablation:numdecoderlayer}
  	\end{subtable}\hspace{2mm}
  	\begin{subtable}{0.32\linewidth}
  	\centering
 		\tablestyle{3pt}{1.2}
 		\scriptsize
  	\begin{tabular}{l | x{15}x{15}x{15}x{15} | x{20}x{15}}
  		$N_a$& \apm$^\texttt{val}$  & \aps & FLOPs & FPS \\
  		\shline
  		10 & 31.2 & 10.2 & 32.9 & 54.2 & 71.9G & 39.8  \\
		50 & 36.9 & 14.5 & 39.6 & 58.5 & 73.5G & 37.7  \\
		100 & 37.9 & 16.0 & 40.7 & 60.1 & 75.5G & 35.5 \\ 
		200 & 38.5 & 16.2 & 41.3 & \textbf{60.8} & 79.5G & 31.6 \\
		300 & \textbf{38.8} & \textbf{17.5} & \textbf{42.0} & 60.2 & 83.5G & 28.6 \\
  	\end{tabular}
	\caption{\textbf{Effect of IA-guided query number on AP.} Increasing IA-guided query number contributes to AP performance.}
	\label{tab:ablation:numiaqueryap}
	\end{subtable}\hspace{2mm}
  	\begin{subtable}{0.32\linewidth}
	\centering
	\tablestyle{3pt}{1.2}
	\scriptsize
	\begin{tabular}{l | x{20}x{18}x{18}x{18}}
		$N_a$& \phantom{0}AR$^\texttt{val}$  & AR$_\text{S}$ & AR$_\text{M}$ & AR$_\text{L}$ \\
		\shline
		10 & \phantom{0}38.2 & 14.4 & 40.5 & 62.8   \\
		50 & \phantom{0}49.8 & 26.2 & 53.3 & 72.6   \\
		100 & \phantom{0}52.0 & 28.9 & 55.3 & 74.2  \\ 
		200 & \phantom{0}53.4 & 30.5 & 57.4 & 74.9  \\
		300 & \phantom{0}\textbf{54.0} & \textbf{31.3} & \textbf{57.6} & \textbf{75.3}  \\
	\end{tabular}
	\caption{\textbf{Effect of IA-guided query number on AR@100.}  Larger IA-guided query number improves object recalls, especially for small objects.}
	\label{tab:ablation:numiaqueryar}
	\end{subtable}
	\vspace{2mm}

 	\begin{subtable}{0.32\linewidth}
	\centering
	\tablestyle{3pt}{1.2}
	\scriptsize
	\begin{tabular}{y{8}| x{10} | x{14}x{14}x{14}x{14} | x{18}x{14}}
		$N_b$ & $N_a$ & \apm & \aps & FLOPs & FPS \\
		\shline
		0 &100 & 37.7 & 15.7 & 40.4 & \textbf{60.3} & 75.2G & 35.6 \\
		0 &108 & 37.6 & 15.6 & 40.7 & 59.9 & 75.6G & 35.3 \\
		8 &100 & \textbf{37.9} & 16.0 & 40.7 & 60.1 & 75.5G & 35.5 \\
		16 & 100 & 37.8 & \textbf{16.1} & \textbf{40.9} & 59.9 & 75.7G & 35.1 \\
	\end{tabular}
	\caption{\textbf{Auxiliary learnable query number.} Adding a few (\ie, 8) auxiliary learnable queries performs better than setting all queries as IA-guided ones.}
	\label{tab:ablation:numauxquery}
	\end{subtable}\hspace{2mm}
	\begin{subtable}{0.3\linewidth}
	\centering
	\tablestyle{3pt}{1.2}
	\scriptsize
	\begin{tabular}{l | x{15}x{15}x{15}x{15} | x{20}x{15}}
		& \apm & \aps & FLOPs & FPS \\
		\shline
		E$_5$ & 37.8 & 15.5 & 40.7 & 60.2 & 74.6G & 35.6 \\
		E$_4$ & 37.9 & 16.0 & 40.7 & 60.1 & 75.5G & 35.5 \\
		E$_3$ & \textbf{38.0} & \textbf{16.3} & \textbf{41.2} & \textbf{60.6} & 79.1G & 34.7 \\
	\end{tabular}
	\caption{\textbf{Query selection source.} Selecting IA-guided queries from larger feature maps leads to better results, but the gain is limited. E$_4$ is a trade-off choice between accuracy and speed.
	}
	\label{tab:ablation:querylevel}
	\end{subtable}\hspace{2mm}
	\begin{subtable}{0.34\linewidth}
	\centering
	\tablestyle{3pt}{1.2}
	\scriptsize
	\begin{tabular}{l | x{13}x{13}x{13}x{13} | x{18}x{14}}
		 & \apm & \aps & FLOPs & FPS \\
		\shline
		Baseline & \textbf{37.9} & \textbf{16.0} & \textbf{40.7} & \textbf{60.1} & 75.5G & 35.5 \\
		$-$ bi. matching & 35.7	& 14.2 & 38.1 & 57.5 & 75.5G & 35.5 \\
		$-$ loc. cost & 37.1 & 14.9 & 40.0 & 59.3 & 75.5G & 35.5  \\
	\end{tabular}
	\caption{\textbf{Instance activation loss.} We remove one component at a time. When removing the bipartite matching strategy, we use a fixed target assignment for each pixel according to their semantic class labels.}
	\label{tab:ablation:ialoss}
	\end{subtable}
	
	\caption{\textbf{Several ablations for \modelname}. Results are evaluated on COCO \texttt{val2017}.}
	\label{tab:ablation:maskformer}
\end{table*}

\begin{table}[t]
	\centering
	\tablestyle{2pt}{1.2}
	\scriptsize
	\begin{tabular}{y{84} |x{20} x{18} x{18}x{18}|x{22}x{18}}
		 & \phantom{0}\apm$^\texttt{val}$ & \aps &  \phantom{0}FLOPs & FPS \\
		\shline
		\modelname-D3 & 37.9 & 16.0 & 40.7 & 60.1 &  \phantom{0}75.5G & 35.5 \\
		$-$ learnable pos. embeddings & 37.9 & 16.4 & 40.6 & \textbf{60.3} & \phantom{0}75.5G & 32.9 \\
		\hline
	\end{tabular}
	\caption{\textbf{Positional embeddings.} When removing  learnable positional embeddings, we use the non-parametric sinusoidal positional embeddings, as in \cite{detr, cheng2021mask2former} }
	\label{tab:ablation:largefeature}
\end{table}

\subsection{Ablation studies}

We now perform a series of ablation studies to analyze \modelname. We first verify the effectiveness of three proposed key components, \ie, IA-guided queries, dual-path update strategy, and GT mask-guided learning, and then study the effect of some other designs about \modelname. Unless specified otherwise, we conduct experiments on \modelname-D3 with ResNet-50~\cite{he2016deep} backbone. All ablation results are evaluated on the COCO \texttt{val2017} set.

\noindent\textbf{IA-guided queries.} As shown in \tabref{tab:ablation:iaquery}, our IA-guided queries achieve better results than zero~\cite{detr} or learning-based ones~\cite{cheng2021mask2former}. Recent work~\cite{pei2022osformer} proposes to use resized multi-scale features as instance queries. However, such a fix-position query selection strategy is hard to extract representative embeddings for all potential objects and thus obtains lower performance. Note that IA-guide queries achieve a more significant result when the model is equipped with only one Transformer decoder layer, which shows their great efficiency in lightweight model design.

\noindent\textbf{Dual-path update strategy.} \tabref{tab:ablation:dualpath} shows the effectiveness of our dual-path update strategy. Thanks to the co-optimization of query and pixel features, our dual-path update strategy performs better than the conventional single query update strategy~\cite{cheng2021mask2former} in our lightweight pixel decoder setting. 
The update order of query and pixel features does not matter much in our experiments.

\noindent\textbf{GT mask-guided learning}. As shown in \tabref{tab:ablation:gtmask}, GT mask-guided learning improves the model performance by up to 0.5 AP, indicating that this technique indeed helps the Transformer decoder learn how to update queries for better object embeddings under masked attention mechanism. \tabref{tab:ablation:gtmask} also demonstrates the generality of GT mask-guided learning for different backbones.

The above ablations demonstrate the effectiveness of our proposed three key techniques. We refer interested readers to the Appendix for more ablations about them, \eg, the changes and the corresponding improvements based on the original Mask2Former.
We then explore the effect of some other designs about \modelname.

\noindent\textbf{Pixel decoder.}
\modelname is compatible with any existing pixel decoders.
\tabref{tab:ablation:pixeldecoder} shows the performance of \modelname with different pixel decoders. Stronger pixel decoders produce better contextual features and lead to higher results but consume more computation. For fast real-time instance segmentation, PPM-FPN~\cite{cheng2022sparseInst} is a good trade-off choice.

\noindent\textbf{Transformer decoder layer number.} As shown in \tabref{tab:ablation:numdecoderlayer}, increasing the number of Transformer decoder layers contributes to the segmentation performance in \modelname. In particular, the mask performance achieves 30.5 AP without using the Transformer decoder. This is mainly attributed to the effectiveness of IA-guided queries, which carry rich embedding information about potential objects at the initial. In addition, our segmentation performance is saturated at around the sixth layer. Continuing to increase decoder layers only marginally improve it.
Also note that FastInst can obtain good performance with only a few Transformer decoder layers,  which is advantageous in real-time.

\begin{figure}[t]
    \begin{subtable}{1.0\linewidth}
    \centering
    \bgroup
    \def\arraystretch{0.2}
    \setlength\tabcolsep{0.2pt}
    \begin{tabular}{cccc}
    \includegraphics[width=0.24\linewidth, height=0.18\linewidth]{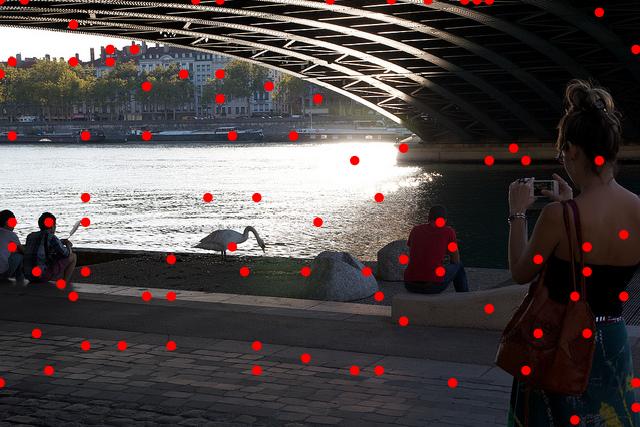} &
    \includegraphics[width=0.24\linewidth, height=0.18\linewidth]{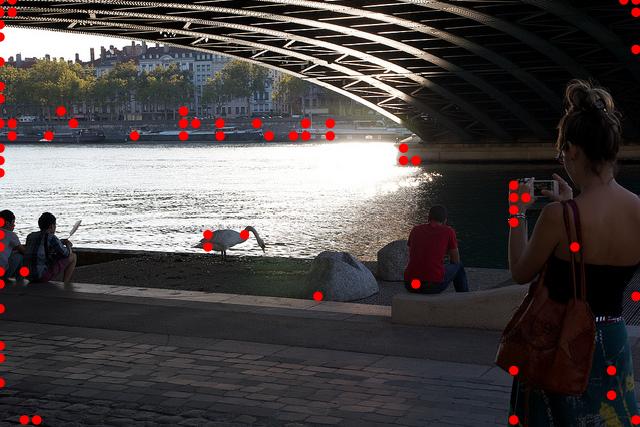} &
    \includegraphics[width=0.24\linewidth, height=0.18\linewidth]{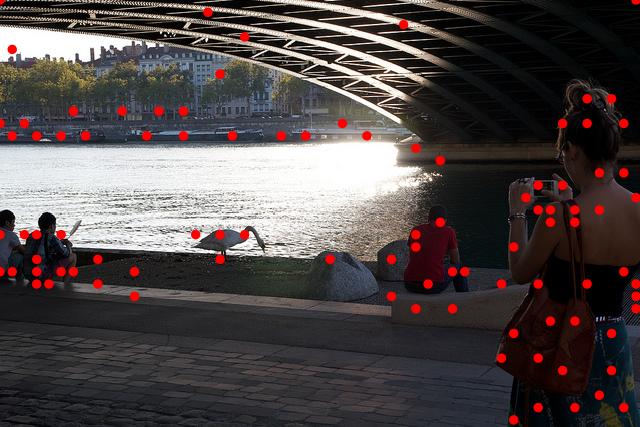} &
    \includegraphics[width=0.24\linewidth, height=0.18\linewidth]{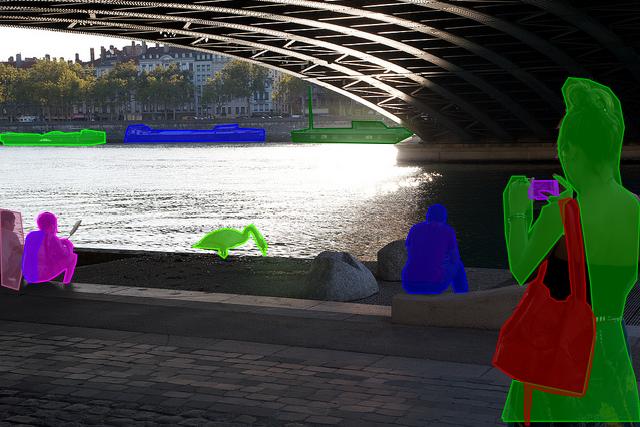} \\
    \includegraphics[width=0.24\linewidth, height=0.18\linewidth]{figures/vis_query_points/img_000000002473/000000002473_semantic.jpg} &
	\includegraphics[width=0.24\linewidth, height=0.18\linewidth]{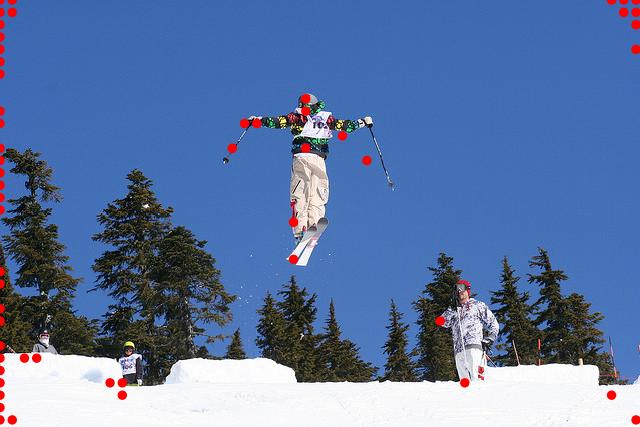} &
	\includegraphics[width=0.24\linewidth, height=0.18\linewidth]{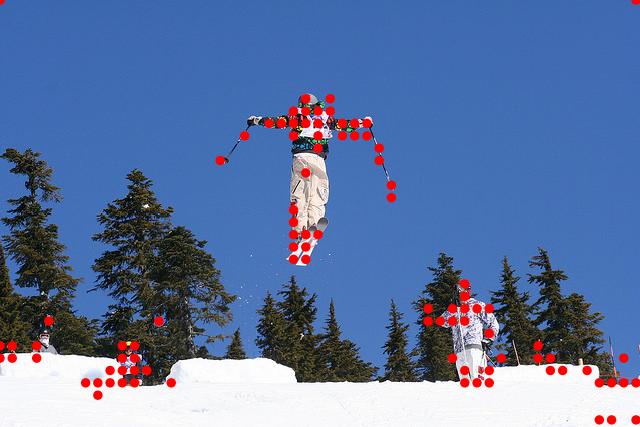} &
	\includegraphics[width=0.24\linewidth, height=0.18\linewidth]{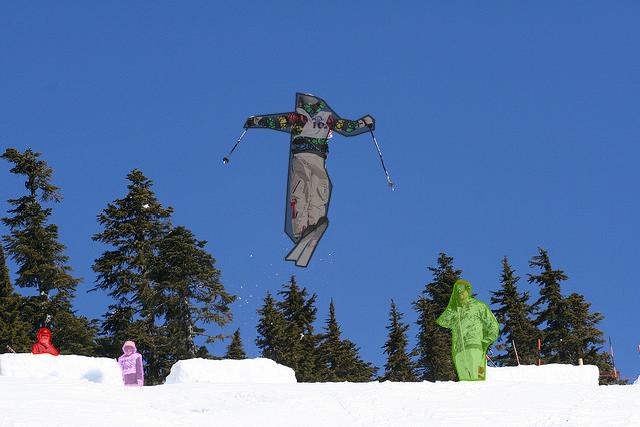} \\
    \includegraphics[width=0.24\linewidth, height=0.18\linewidth]{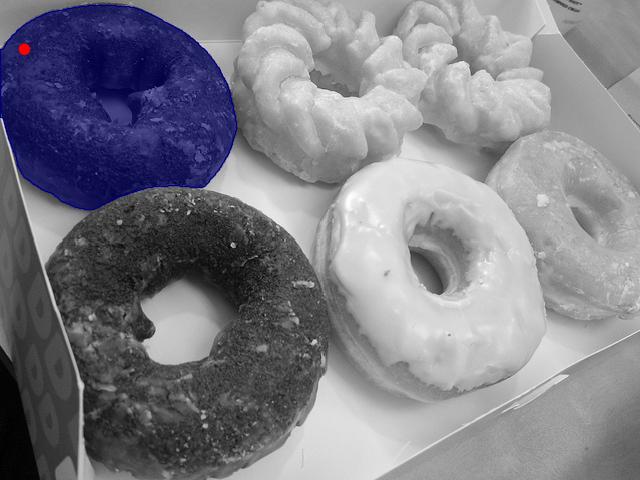} &
	\includegraphics[width=0.24\linewidth, height=0.18\linewidth]{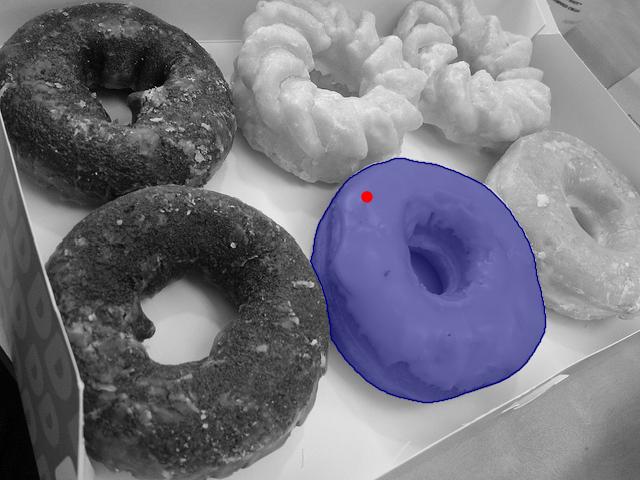} &
	\includegraphics[width=0.24\linewidth, height=0.18\linewidth]{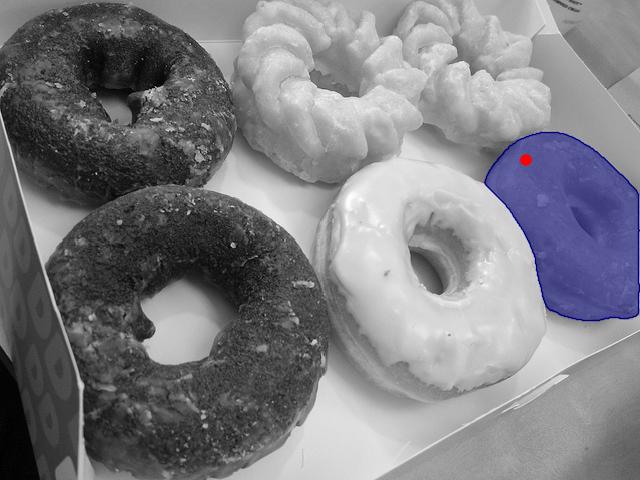} &
	\includegraphics[width=0.24\linewidth, height=0.18\linewidth]{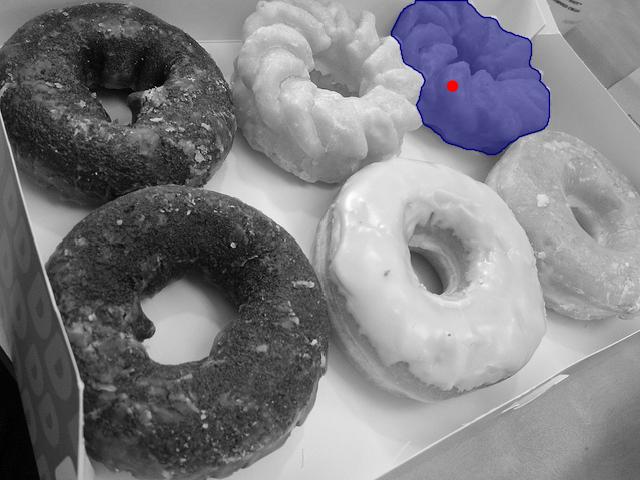} \\
    \end{tabular} \egroup
    \end{subtable}
  \caption{\textbf{Visualization of IA-guided queries.} The \textbf{first and second rows} show the distributions of IA-guided queries with different supervision losses for the auxiliary classification head. First column: dense pixel-wise semantic classification loss. Second column: bipartite matching-based Hungarian loss without the location cost. Third column: bipartite matching-based Hungarian loss with the location cost (ours). Fourth column: ground truth. The query points under our designed loss (third column) are more concentrated on the region of each foreground object. The \textbf{last row} shows four predicted masks (\textbf{blue}) with corresponding IA-guided query locations (\textbf{red}).}
  \label{fig:analysis:query}
\end{figure}

\noindent\textbf{IA-guided query number.} In Mask2Former, increasing the query number to more than 100 will slightly degrade the instance segmentation performance. In \tabref{tab:ablation:numiaqueryap}, the results indicate that increasing the number of IA-guided queries will contribute to the segmentation performance in \modelname. We attribute this to the improved object recall (see \tabref{tab:ablation:numiaqueryar}) and increased object embedding information for decoding. On the other hand, growing IA-guided queries will affect the model inference speed. Note that even with 10 IA-guided queries, our model can obtain 31.2 AP on COCO dataset, which has 7.7 instances per image on average~\cite{lin2014coco}. This indicates the effectiveness of IA-guided queries again.

\noindent\textbf{Auxiliary learnable query number.} Auxiliary queries aim to gather background and image-independent information for pixel feature updates and query updates. They do not participate in object predictions. \tabref{tab:ablation:numauxquery} shows that adding a few auxiliary learnable queries helps improve the performance, better than setting all queries as IA-guided queries.

\noindent\textbf{Query selection source.} As shown in \tabref{tab:ablation:querylevel}, selecting IA-guided queries from larger feature maps leads to better results. E$_4$ is a good trade-off choice between accuracy and speed. Nevertheless, the contribution of the selection source to the model performance is limited.

\noindent\textbf{Instance activation loss.} We study the effect of two components in instance activation loss. 
As shown in \tabref{tab:ablation:ialoss}, the bipartite matching-based target assignment strategy leads to a significant gain, which provides a sparse pixel embedding activation for IA-guided query selection. 
Here when removing the bipartite matching strategy, we use the semantic class label as the target of each pixel, as in common semantic segmentation tasks~\cite{deeplabV2,deeplabV3plus}. 
The location cost also plays a vital role in the matching loss, which reduces matching space and accelerates model convergence. 
\figref{fig:analysis:query} visualizes the distributions of  IA-guided queries, which also shows the superiority of our designed loss.

\noindent\textbf{Positional embeddings.}  \tabref{tab:ablation:largefeature} demonstrates that using learnable positional embeddings instead of non-parametric sinusoidal positional embeddings can improve the model inference speed without compromising the performance.


\section{Conclusion}
We propose \modelname for real-time instance segmentation. Built on a query-based segmentation framework~\cite{cheng2021mask2former} and three designed efficient components, \ie, instance activation-guided queries, dual-path update strategy, and ground truth mask-guided learning, \modelname achieves excellent performance on the popular COCO dataset while maintaining a fast inference speed. Extensive experiments demonstrate the effectiveness of core ideas and the superiority of \modelname over previous state-of-the-art real-time counterparts. We hope this work can serve as a new baseline for real-time instance segmentation and promote the development of query-based image segmentation algorithms. 

\noindent\textbf{Limitations.} (1) Like general query-based models~\cite{detr, cheng2021mask2former,li2021panopticsegformer}, \modelname is not good at small targets. Even though using stronger pixel decoders or larger feature maps improves it, it introduces heavier computational burdens, and the result is still unsatisfactory. We look forward to an essential solution to handle this problem. (2) although GT mask-guided learning improves the performance of masked attention, it increases training costs. We hope a more elegant method can be proposed to replace it.

{\small
\bibliographystyle{ieee_fullname}
\bibliography{egbib}

\begin{thebibliography}{10}\itemsep=-1pt

\bibitem{yolact-iccv2019}
Daniel {Bolya}, Chong {Zhou}, Fanyi {Xiao}, and Yong~Jae {Lee}.
\newblock Yolact: {Real-time} instance segmentation.
\newblock In {\em ICCV}, 2019.

\bibitem{yolact-plus-tpami2020}
Daniel {Bolya}, Chong {Zhou}, Fanyi {Xiao}, and Yong~Jae {Lee}.
\newblock Yolact++: Better real-time instance segmentation.
\newblock {\em PAMI}, 2020.

\bibitem{cai2018cascade}
Zhaowei Cai and Nuno Vasconcelos.
\newblock {Cascade R-CNN}: Delving into high quality object detection.
\newblock In {\em CVPR}, 2018.

\bibitem{detr}
Nicolas Carion, Francisco Massa, Gabriel Synnaeve, Nicolas Usunier, Alexander
  Kirillov, and Sergey Zagoruyko.
\newblock End-to-end object detection with transformers.
\newblock In {\em ECCV}, 2020.

\bibitem{chen2019hybrid}
Kai Chen, Jiangmiao Pang, Jiaqi Wang, Yu Xiong, Xiaoxiao Li, Shuyang Sun,
  Wansen Feng, Ziwei Liu, Jianping Shi, Wanli Ouyang, et~al.
\newblock Hybrid task cascade for instance segmentation.
\newblock In {\em CVPR}, 2019.

\bibitem{deeplabV2}
Liang-Chieh Chen, George Papandreou, Iasonas Kokkinos, Kevin Murphy, and Alan~L
  Yuille.
\newblock {DeepLab}: Semantic image segmentation with deep convolutional nets,
  atrous convolution, and fully connected {CRF}s.
\newblock {\em PAMI}, 2018.

\bibitem{deeplabV3plus}
Liang-Chieh Chen, Yukun Zhu, George Papandreou, Florian Schroff, and Hartwig
  Adam.
\newblock Encoder-decoder with atrous separable convolution for semantic image
  segmentation.
\newblock In {\em ECCV}, 2018.

\bibitem{chen2019tensormask}
Xinlei Chen, Ross Girshick, Kaiming He, and Piotr Doll{\'a}r.
\newblock {TensorMask}: A foundation for dense object segmentation.
\newblock In {\em ICCV}, 2019.

\bibitem{cheng2021mask2former}
Bowen {Cheng}, Ishan {Misra}, Alexander~G. {Schwing}, Alexander {Kirillov}, and
  Rohit {Girdhar}.
\newblock Masked-attention mask transformer for universal image segmentation.
\newblock In {\em CVPR}, 2022.

\bibitem{cheng2021maskformer}
Bowen Cheng, Alexander~G. Schwing, and Alexander Kirillov.
\newblock Per-pixel classification is not all you need for semantic
  segmentation.
\newblock In {\em NeurIPS}, 2021.

\bibitem{cheng2022sparseInst}
Tianheng Cheng, Xinggang Wang, Shaoyu Chen, Wenqiang Zhang, Qian Zhang, Chang
  Huang, Zhaoxiang Zhang, and Wenyu Liu.
\newblock Sparse instance activation for real-time instance segmentation.
\newblock In {\em CVPR}, 2022.

\bibitem{chengwhl20}
Tianheng {Cheng}, Xinggang {Wang}, Lichao {Huang}, and Wenyu {Liu}.
\newblock Boundary-preserving mask r-cnn.
\newblock In {\em ECCV}, 2020.

\bibitem{Cordts2016Cityscapes}
Marius Cordts, Mohamed Omran, Sebastian Ramos, Timo Rehfeld, Markus Enzweiler,
  Rodrigo Benenson, Uwe Franke, Stefan Roth, and Bernt Schiele.
\newblock The {Cityscapes} dataset for semantic urban scene understanding.
\newblock In {\em CVPR}, 2016.

\bibitem{dong2021solq}
Bin Dong, Fangao Zeng, Tiancai Wang, Xiangyu Zhang, and Yichen Wei.
\newblock Solq: Segmenting objects by learning queries.
\newblock In {\em NeurIPS}, 2021.

\bibitem{du2021realtime}
Wentao Du, Zhiyu Xiang, Shuya Chen, Chengyu Qiao, Yiman Chen, and Tingming Bai.
\newblock Real-time instance segmentation with discriminative orientation maps.
\newblock In {\em ICCV}, 2021.

\bibitem{he2017mask}
Kaiming He, Georgia Gkioxari, Piotr Doll{\'a}r, and Ross Girshick.
\newblock Mask {R-CNN}.
\newblock In {\em ICCV}, 2017.

\bibitem{he2016deep}
Kaiming He, Xiangyu Zhang, Shaoqing Ren, and Jian Sun.
\newblock Deep residual learning for image recognition.
\newblock In {\em CVPR}, 2016.

\bibitem{he2019bag}
Tong He, Zhi Zhang, Hang Zhang, Zhongyue Zhang, Junyuan Xie, and Mu Li.
\newblock Bag of tricks for image classification with convolutional neural
  networks.
\newblock In {\em CVPR}, 2019.

\bibitem{hu2021istr}
Jie Hu, Liujuan Cao, Yao Lu, ShengChuan Zhang, Yan Wang, Ke Li, Feiyue Huang,
  Ling Shao, and Rongrong Ji.
\newblock Istr: End-to-end instance segmentation with transformers.
\newblock {\em arXiv:2105.00637}, 2021.

\bibitem{huang2019mask}
Zhaojin Huang, Lichao Huang, Yongchao Gong, Chang Huang, and Xinggang Wang.
\newblock Mask scoring {R-CNN}.
\newblock In {\em CVPR}, 2019.

\bibitem{kirillov2019panopticfpn}
Alexander Kirillov, Ross Girshick, Kaiming He, and Piotr Doll{\'a}r.
\newblock Panoptic feature pyramid networks.
\newblock In {\em CVPR}, 2019.

\bibitem{kirillov2020pointrend}
Alexander Kirillov, Yuxin Wu, Kaiming He, and Ross Girshick.
\newblock {PointRend}: Image segmentation as rendering.
\newblock In {\em CVPR}, 2020.

\bibitem{kuhn1955hungarian}
Harold~W Kuhn.
\newblock The hungarian method for the assignment problem.
\newblock {\em Naval research logistics quarterly}, 1955.

\bibitem{lee2020centermask}
Youngwan Lee and Jongyoul Park.
\newblock Centermask: Real-time anchor-free instance segmentation.
\newblock In {\em CVPR}, 2020.

\bibitem{li2022mask}
Feng {Li}, Hao {Zhang}, Huaizhe {xu}, Shilong {Liu}, Lei {Zhang}, Lionel~M.
  {Ni}, and Heung-Yeung {Shum}.
\newblock Mask dino: Towards a unified transformer-based framework for object
  detection and segmentation.
\newblock {\em arXiv:2206.02777}, 2022.

\bibitem{li2021panopticsegformer}
Zhiqi Li, Wenhai Wang, Enze Xie, Zhiding Yu, Anima Anandkumar, Jose~M. Alvarez,
  Ping Luo, and Tong Lu.
\newblock Panoptic segformer: Delving deeper into panoptic segmentation with
  transformers.
\newblock In {\em CVPR}, 2022.

\bibitem{lin2016feature}
Tsung-Yi Lin, Piotr Doll{\'a}r, Ross Girshick, Kaiming He, Bharath Hariharan,
  and Serge Belongie.
\newblock Feature pyramid networks for object detection.
\newblock In {\em CVPR}, 2017.

\bibitem{lin2014coco}
Tsung-Yi Lin, Michael Maire, Serge Belongie, James Hays, Pietro Perona, Deva
  Ramanan, Piotr Doll{\'a}r, and C~Lawrence Zitnick.
\newblock Microsoft {COCO}: Common objects in context.
\newblock In {\em ECCV}, 2014.

\bibitem{liu2018path}
Shu Liu, Lu Qi, Haifang Qin, Jianping Shi, and Jiaya Jia.
\newblock Path aggregation network for instance segmentation.
\newblock In {\em CVPR}, 2018.

\bibitem{long2015fully}
Jonathan Long, Evan Shelhamer, and Trevor Darrell.
\newblock Fully convolutional networks for semantic segmentation.
\newblock In {\em CVPR}, 2015.

\bibitem{loshchilov2018decoupled}
Ilya Loshchilov and Frank Hutter.
\newblock Decoupled weight decay regularization.
\newblock In {\em ICLR}, 2019.

\bibitem{milletari2016v}
Fausto Milletari, Nassir Navab, and Seyed-Ahmad Ahmadi.
\newblock {V-Net}: Fully convolutional neural networks for volumetric medical
  image segmentation.
\newblock In {\em 3DV}, 2016.

\bibitem{pei2022osformer}
Jialun Pei, Tianyang Cheng, Deng-Ping Fan, He Tang, Chuanbo Chen, and Luc
  Van~Gool.
\newblock Osformer: One-stage camouflaged instance segmentation with
  transformers.
\newblock In {\em ECCV}, 2022.

\bibitem{redmon2018yolov3}
Joseph {Redmon} and Ali {Farhadi}.
\newblock {YOLOv3: An Incremental Improvement}.
\newblock {\em arXiv:1804.02767}, 2018.

\bibitem{Ren2015a}
Shaoqing Ren, Kaiming He, Ross Girshick, and Jian Sun.
\newblock {Faster R-CNN}: Towards real-time object detection with region
  proposal networks.
\newblock In {\em NeurIPS}, 2015.

\bibitem{stewart2016}
Russell {Stewart} and Mykhaylo {Andriluka}.
\newblock {End-to-end people detection in crowded scenes}.
\newblock In {\em CVPR}, 2016.

\bibitem{tang2021look}
Chufeng {Tang}, Hang {Chen}, Xiao {Li}, Jianmin {Li}, Zhaoxiang {Zhang}, and
  Xiaolin {Hu}.
\newblock Look closer to segment better: Boundary patch refinement for instance
  segmentation.
\newblock In {\em CVPR}, 2021.

\bibitem{tian2020conditional}
Zhi Tian, Chunhua Shen, and Hao Chen.
\newblock Conditional convolutions for instance segmentation.
\newblock In {\em ECCV}, 2020.

\bibitem{tian2021fcos}
Zhi Tian, Chunhua Shen, Hao Chen, and Tong He.
\newblock {FCOS}: A simple and strong anchor-free object detector.
\newblock {\em PAMI}, 2021.

\bibitem{vaswani2017attention}
Ashish Vaswani, Noam Shazeer, Niki Parmar, Jakob Uszkoreit, Llion Jones,
  Aidan~N Gomez, Lukasz Kaiser, and Illia Polosukhin.
\newblock Attention is all you need.
\newblock In {\em NeurIPS}, 2017.

\bibitem{wang2021max}
Huiyu Wang, Yukun Zhu, Hartwig Adam, Alan Yuille, and Liang-Chieh Chen.
\newblock {MaX-DeepLab}: End-to-end panoptic segmentation with mask
  transformers.
\newblock In {\em CVPR}, 2021.

\bibitem{wang2020solo}
Xinlong Wang, Tao Kong, Chunhua Shen, Yuning Jiang, and Lei Li.
\newblock {SOLO}: Segmenting objects by locations.
\newblock In {\em ECCV}, 2020.

\bibitem{wang2020solov2}
Xinlong Wang, Rufeng Zhang, Tao Kong, Lei Li, and Chunhua Shen.
\newblock {SOLOv2}: Dynamic and fast instance segmentation.
\newblock {\em NeurIPS}, 2020.

\bibitem{wu2022efficient}
Jialian Wu, Sudhir Yarram, Hui Liang, Tian Lan, Junsong Yuan, Jayan Eledath,
  and Gerard Medioni.
\newblock Efficient video instance segmentation via tracklet query and
  proposal.
\newblock In {\em CVPR}, 2022.

\bibitem{wu2019detectron2}
Yuxin Wu, Alexander Kirillov, Francisco Massa, Wan-Yen Lo, and Ross Girshick.
\newblock Detectron2.
\newblock \url{https://github.com/facebookresearch/detectron2}, 2019.

\bibitem{yu2022cmt}
Qihang {Yu}, Huiyu {Wang}, Dahun {Kim}, Siyuan {Qiao}, Maxwell {Collins}, Yukun
  {Zhu}, Hartwig {Adam}, Alan {Yuille}, and Liang-Chieh {Chen}.
\newblock Cmt-deeplab: Clustering mask transformers for panoptic segmentation.
\newblock In {\em CVPR}, 2022.

\bibitem{zhang2020MEInst}
Rufeng Zhang, Zhi Tian, Chunhua Shen, Mingyu You, and Youliang Yan.
\newblock Mask encoding for single shot instance segmentation.
\newblock In {\em CVPR}, 2020.

\bibitem{zhang2022e2ec}
Tao Zhang, Shiqing Wei, and Shunping Ji.
\newblock E2ec: An end-to-end contour-based method for high-quality high-speed
  instance segmentation.
\newblock In {\em CVPR}, 2022.

\bibitem{zhao2017pspnet}
Hengshuang Zhao, Jianping Shi, Xiaojuan Qi, Xiaogang Wang, and Jiaya Jia.
\newblock Pyramid scene parsing network.
\newblock In {\em CVPR}, 2017.

\bibitem{zhu2019deformable}
Xizhou Zhu, Han Hu, Stephen Lin, and Jifeng Dai.
\newblock Deformable convnets v2: More deformable, better results.
\newblock In {\em CVPR}, 2019.

\bibitem{zhu2021deformable}
Xizhou Zhu, Weijie Su, Lewei Lu, Bin Li, Xiaogang Wang, and Jifeng Dai.
\newblock Deformable detr: Deformable transformers for end-to-end object
  detection.
\newblock In {\em ICLR}, 2021.

\end{thebibliography}
}

\clearpage
\appendix
\begin{center}{\bf \Large Appendix}\end{center}\vspace{-2mm}
\renewcommand{\thetable}{\Roman{table}}
\renewcommand{\thefigure}{\Roman{figure}}
\setcounter{table}{0}
\setcounter{figure}{0}
We first provide several additional ablation studies for \modelname in \appref{app:ablations}. Then, we demonstrate the performance of \modelname on a different dataset, \ie, Cityscapes~\cite{Cordts2016Cityscapes}, in \appref{app:datasets}, and show its unified segmentation ability in \appref{app:unified}. Finally, we visualize many predictions of \modelname on the COCO~\cite{lin2014coco} validation set in \appref{app:vis}.

\section{Additional ablation studies}
\label{app:ablations}

\subsection{Improvements on original Mask2Former}

Considering that the proposed \modelname is developed based on Mask2Former, we investigate the improvements of three proposed key components, \ie, instance activation-guided queries, dual-path update strategy, and ground truth mask-guided learning, on the original Mask2Former. The results are shown in \tabref{tab:ablation:mask2former}. Dual-path update strategy improves Mask2Former's efficiency the most, not only accelerating its inference speed but also reducing the model parameters dramatically. GT-mask guided learning also performs well on the original Mask2Former. The instance activation-guided queries improve the original Mask2Former little since the original Mask2Former already has enough Transformer decoder layers (\ie, nine layers), and the learnable queries can also be decoded well. Note that IA-guided queries contribute to the 3-layer dual-path Transformer decoder.

\subsection{Effect of location cost on deformable convolutional networks}
We employ Hungarian loss~\cite{detr} with a location cost during training to supervise the auxiliary classification head.
The location cost restricts the matched pixels inside the object region,  which reduces the matching space and, thus, accelerates training convergence.
\tabref{tab:ablation:loc} shows that this location cost is also effective for \modelname with the backbone that employs deformable convolutional networks (DCNs)~\cite{zhu2019deformable}. DCNs add 2D offsets to the regular grid sampling locations in the standard convolution and enable the pixels not located in the object region to have a chance of being activated for the segmentation. Despite this, the pixels outside the object region are not good IA-guided query candidates since they rely on precise offset predictions.
\figref{fig:vis_dcn_query_points} visualizes the distributions of IA-guided queries with/without the location cost in a FastInst-D1 model with DCNs. The location cost helps produce more concentrated and higher-quality IA-guided queries.

\subsection{Effect of local-maximum-first selection strategy}
During prediction, we first select the pixel embeddings in E$_4$ with $p_{i,k_i}$ that is the local maximum in the corresponding class plane and then pick the ones with the top foreground probabilities. Such a local-maximum-first selection strategy prevents the selected IA-guided queries from focusing on some salient objects.
\tabref{tab:ablation:ms} demonstrates the effectiveness of the local-maximum-first selection strategy. It improves the performance, especially when the IA-guided query number is small.
\figref{fig:vis_q10_query_points} shows the influence of the local-maximum-first selection strategy on the selected IA-guided queries. Without the local-maximum-first selection strategy, two selected queries fall on the same salient object (\ie, handbag), which hurts the recall of other instances.

\begin{table}[t]
	\centering
	\scriptsize
  \setlength\tabcolsep{1.7mm}
	\begin{tabular}{lccccccccccc}
	A & & & & & \checkmark & \checkmark & \checkmark & \checkmark & \checkmark\\
   B & & \checkmark & & \checkmark & & \checkmark & & \checkmark & \checkmark\\
   C & & & \checkmark & \checkmark & & & \checkmark & \checkmark & \checkmark\\
   E & & & & & & & & & \checkmark \\
   \hline
   Param.(M) & 40.9 & 41.4 & 40.9 & 41.4 & \textbf{33.5} & 34.1 & \textbf{33.5} & 34.1 & 34.2 \\
   FPS & 25.3 & 24.5 & 25.3 & 24.5 & 34.3 & 32.9 & 34.3 & 32.9 & \textbf{35.5} \\
   AP$^\texttt{val}_{coco}$ & 37.2 & 37.3 & 37.6 & 37.6 & 36.9 & 37.3 & 37.5 & 37.8 & \textbf{37.9} \\
	\end{tabular}
	\caption{\textbf{Improvements on original Mask2Former.} A: 3-layer dual-path Transformer decoder (including little head change). B: IA-guided queries. C: GT mask-guided learning. E: learnable positional embeddings and auxiliary queries. The baseline is Mask2Former with PPM-FPN and 9 Transformer decoder layers. With A, B, C, and E, we achieve our FastInst-D3 model (R50 backbone).}
   \label{tab:ablation:mask2former}
\end{table}

\begin{table}[t]
	\centering
	\tablestyle{2pt}{1.2}
	\scriptsize
	\begin{tabular}{p{53px} | x{42} | x{25} x{20} x{20}x{20}|x{23}}
		& backbone & \phantom{0}\apm$^\texttt{val}$ & \aps & FPS \\
		\shline
		w/o location cost & R50-d-DCN & 36.5 & 14.8 & 39.6 & 59.2 & 43.3 \\
		\textbf{w/  location cost} & R50-d-DCN & \textbf{38.1}\tiny(\gain{+1.6}) & \textbf{16.2} & \textbf{41.5} & \textbf{60.6} & 43.3 \\
		\hline
	\end{tabular}
	\caption{\textbf{Effect of location cost on DCNs.} The location cost is also important for \modelname with the backbone that employs DCNs. We conduct ablation studies on the \modelname-D1-640 model.}
	\vspace{-3mm}
	\label{tab:ablation:loc}
\end{table}

\begin{table}[t]
	\centering
	\tablestyle{2pt}{1.2}
	\scriptsize
	\begin{tabular}{p{75px} | x{20} | x{25} x{20} x{20}x{20}|x{23}}
		& $N_a$ & \phantom{0}\apm$^\texttt{val}$ & \aps & FPS \\
		\shline
		w/o local-maximum-first & 10 & 28.6 & 9.0 & 30.0 & 49.3 & 53.5\\
		\textbf{w/  local-maximum-first} & 10 & \textbf{30.0}\tiny(\gain{+1.4}) & \textbf{9.8} & \textbf{32.1} & \textbf{51.3} & 52.9 \\
		\hline\hline
		w/o local-maximum-first & 50 & 34.4 & 13.9 & 36.9 & 55.3 & 51.3 \\
		\textbf{w/  local-maximum-first} & 50 & \textbf{34.8}\tiny(\gain{+0.4}) & \textbf{14.2} & \textbf{37.6} & \textbf{55.7} & 51.0 \\
    \hline\hline
		w/o local-maximum-first & 100 & 35.3 & \textbf{14.6} & 38.2 & 56.3 & 49.0 \\
		\textbf{w/  local-maximum-first} & 100 & \textbf{35.6}\tiny(\gain{+0.3}) & 14.3 & \textbf{38.8} & \textbf{56.6} & 48.8 \\
    \hline
	\end{tabular}
	\caption{\textbf{Effect of local-maximum-first selection strategy.} The local-maximum-first selection strategy is effective, especially when the IA-guided query number (\ie, $N_a$) is small. We conduct ablation studies on \modelname-D1-640 with ResNet-50 backbone.}
	\label{tab:ablation:ms}
\end{table}

\begin{figure}[!t]
  \centering
  \begin{adjustbox}{width=\linewidth}
  \bgroup
  \def\arraystretch{0.2}
  \setlength\tabcolsep{0.2pt}
  \begin{tabular}{cc}
  \includegraphics[width=0.49\linewidth]{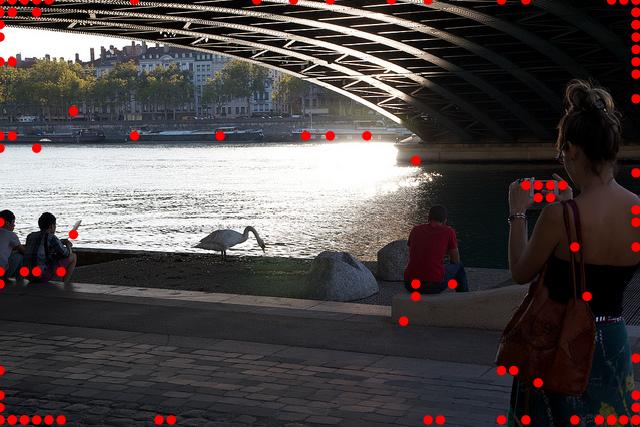} &
  \includegraphics[width=0.49\linewidth]{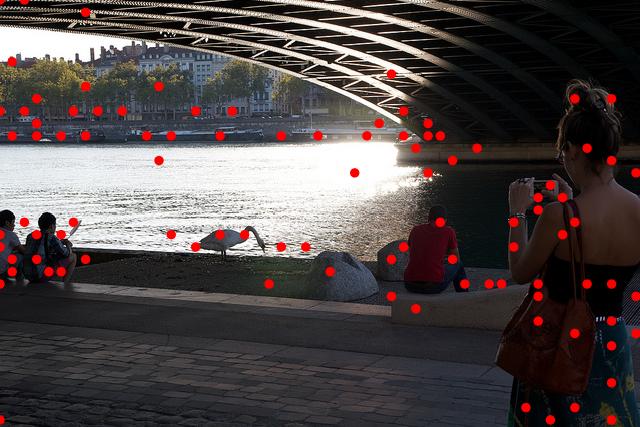} \\
  \end{tabular} \egroup
  \end{adjustbox}
  \caption{
    \textbf{Effect of location cost on IA-guided queries with DCNs.} Left: not use the location cost. Right: use the location cost. We visualize the distributions of IA-guided queries in \modelname-D1-640 with a ResNet-50-d-DCN backbone. A few IA-guided queries are located in the padding area (for the size divisibility of 32) and not shown in the figure.
}
\label{fig:vis_dcn_query_points}
\end{figure}

\begin{figure}[!t]
  \centering
  \begin{adjustbox}{width=\linewidth}
  \bgroup
  \def\arraystretch{0.2}
  \setlength\tabcolsep{0.2pt}
  \begin{tabular}{ccc}
  \includegraphics[width=0.33\linewidth]{figures/visualization/ablations/q10_without_ms/000000001268.jpg} &
  \includegraphics[width=0.33\linewidth]{figures/visualization/ablations/q10_with_ms/000000001268.jpg} &
  \includegraphics[width=0.33\linewidth]{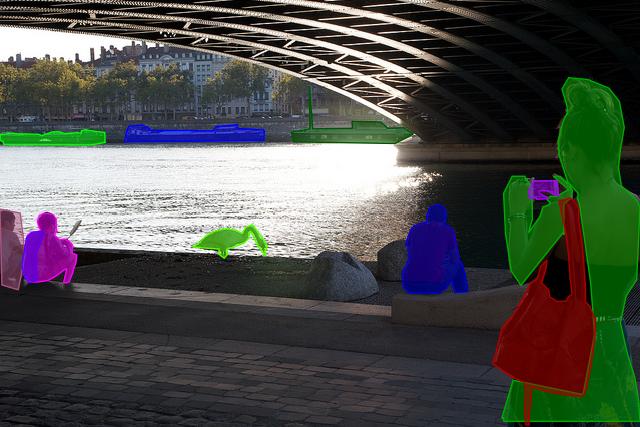} \\
  \end{tabular} \egroup
  \end{adjustbox}
  \caption{
    \textbf{Effect of local-maximum-first selection strategy on IA-guided queries.} We visualize the IA-guided query distributions in \modelname-D1 without (left) and with (middle) the local-maximum-first selection strategy. Here $N_a=10$. The right figure shows the ground truth.
}
\label{fig:vis_q10_query_points}
\end{figure}

\subsection{Auxiliary query analysis}

\begin{figure}
  \centering
  \includegraphics[width=1.0\linewidth]{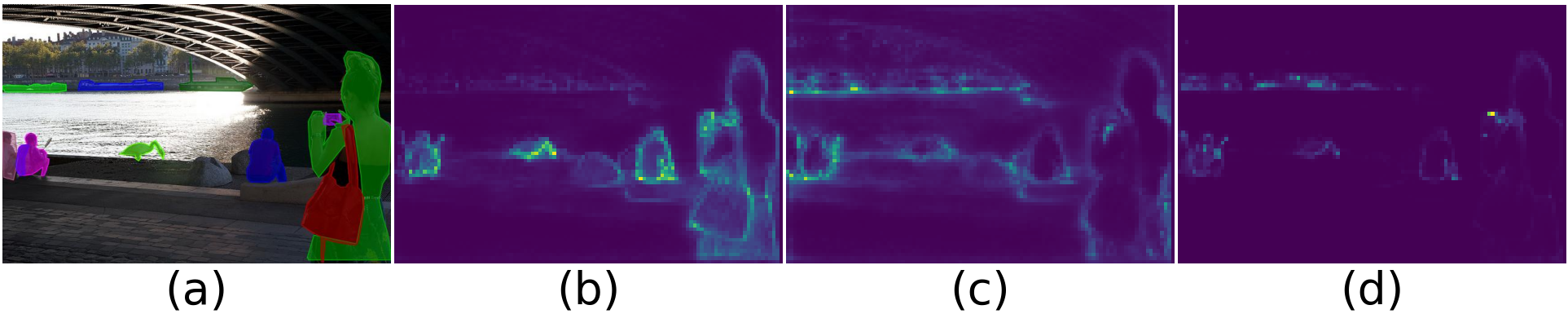}
  \caption{\textbf{Visualization of cross-attention maps of auxiliary learnable queries.} (a) Ground truth. (b-d) Cross-attention maps of three (of eight) auxiliary learnable queries in the last Transformer decoder layer of FastInst-D3. The auxiliary queries in (b) and (c) attend to general edges, including background edges. The auxiliary query in (d) attends to class-agnostic foreground objects.}
  \label{fig:vis_aux_query}
\end{figure}

We visualize the cross-attention maps of three auxiliary learnable queries in the last Transformer decoder layer of FastInst-D3 (R50 backbone) in \figref{fig:vis_aux_query}. As seen, auxiliary queries attend to general information such as edges (including background edge) and class-agnostic foreground objects.

\section{Additional datasets}
\label{app:datasets}
\subsection{Cityscapes}
Cityscapes~\cite{Cordts2016Cityscapes} is a high-resolution ($1024\!\times\!2048$ pixels) street-view dataset that contains 2975 training, 500 validation, and 1525 testing images. We evaluate the performance of FastInst in terms of instance segmentation AP over eight semantic classes of the dataset.

\noindent\textbf{Training settings.}
We use a batch size of 16 and train the model for 90K iterations. We set the initial learning rate as 0.0001 and drop it by multiplying 0.1 at 0.9 and 0.95 fractions of the total number of training steps. During training, we randomly resize the image to a shorter edge from 800 to 1024 pixels with a step of 32 pixels, followed by a crop size of $512\!\times\!1024$. During inference, we operate on the full image with a resolution of $1024\!\times\!2048$.

\noindent\textbf{Results.} \tabref{tab:ablation:cityscapes} shows the result of \modelname on Cityscapes \texttt{val} set. We also report the result of Mask2Former~\cite{cheng2021mask2former} that uses the same pixel decoder, \ie, the pyramid pooling module~\cite{zhao2017pspnet}-based FPN (PPM-FPN) and the same training settings. \modelname outperforms the Mask2Former by a large margin (\ie, 4.1 AP) with a similar speed, showing good efficiency in instance segmentation tasks.

\begin{table}[t]
	\centering
	\tablestyle{2pt}{1.2}
	\scriptsize
	\begin{tabular}{p{60px} | x{40} | x{35} x{30} |x{35}}
		& backbone & \phantom{0}\apm$^\texttt{val}$ & AP$_\text{50}$ & FPS \\
		\shline
		Mask2Former$^{\text{\textdagger}}$ & R50 & 31.4 & 55.9 & 9.2 \\
    \hline
		\textbf{\modelname-D3} (ours) & R50 & \textbf{35.5}\tiny(\gain{+4.1}) & \textbf{59.0} & 9.2 \\
    \hline
	\end{tabular}
	\caption{\textbf{Instance segmentation results on Cityscapes \texttt{val}.} Mask2Former$^{\text{\textdagger}}$ denotes a light version of Mask2Former~\cite{cheng2021mask2former} that uses the same pixel decoder and training settings as \modelname.}
	\label{tab:ablation:cityscapes}
\end{table}

\begin{table}[t]
	\centering
	\scriptsize
	\begin{tabular}{l|c|ccc|cc}
   \multirow{2}{*}{} & \multirow{2}{*}{backbone} & \multicolumn{3}{c|}{Cityscapes \texttt{val}} & \multirow{2}{*}{\#Param. (M)} \\
   & & AP & PQ & mIoU & \\
   \shline
   Mask2Former$^{\text{\textdagger}}$ & R50 & 31.4 & 53.9 & 74.4 & 40.9 \\
   \textbf{FastInst-D3} & R50 & \textbf{35.5} & \textbf{56.4} & \textbf{74.7} & \textbf{34.2} \\
	\end{tabular}
	\caption{\textbf{Panoptic (PQ) and semantic (mIoU) segmentation results on Cityscapes \texttt{val}.} Model and training settings are the same as instance segmentation (see Appendix \ref{app:datasets}). FastInst performs better in instance-level segmentation than Mask2Former.}
   \label{tab:ablation:unified}
\end{table}

\section{Unified Segmentation}
\label{app:unified}

According to the practice of Mask2Former, FastInst can be easily transferred to other segmentation tasks. We show the panoptic and semantic segmentation results on Cityscapes in \tabref{tab:ablation:unified}.

\section{Visualization}
\label{app:vis}
\begin{figure*}[!t]
    \centering
    \begin{adjustbox}{width=\textwidth}
    \bgroup
    \def\arraystretch{0.2}
    \setlength\tabcolsep{0.2pt}
    \begin{tabular}{cccc}
    \includegraphics[height=2cm]{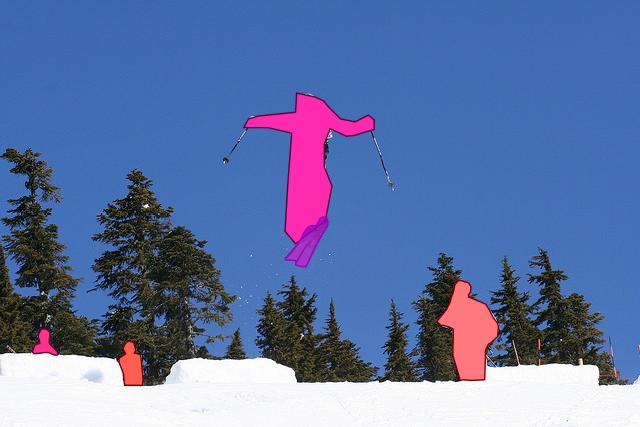} &
    \includegraphics[height=2cm]{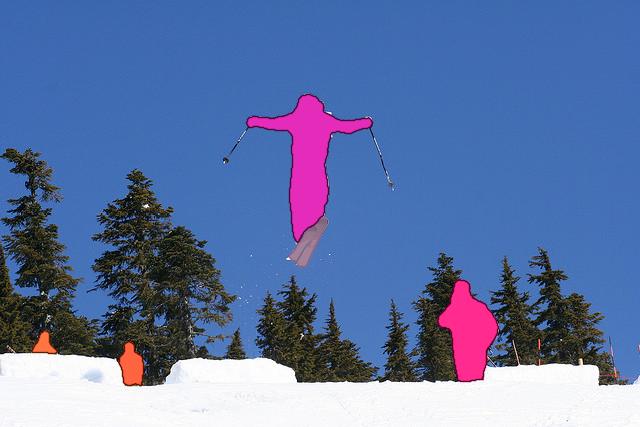} &
    \includegraphics[height=2cm]{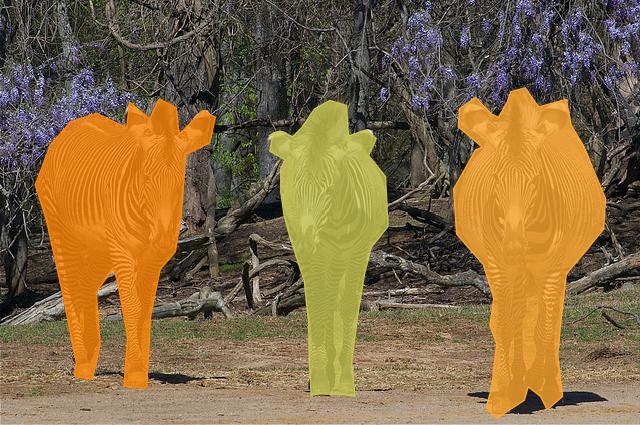} &
    \includegraphics[height=2cm]{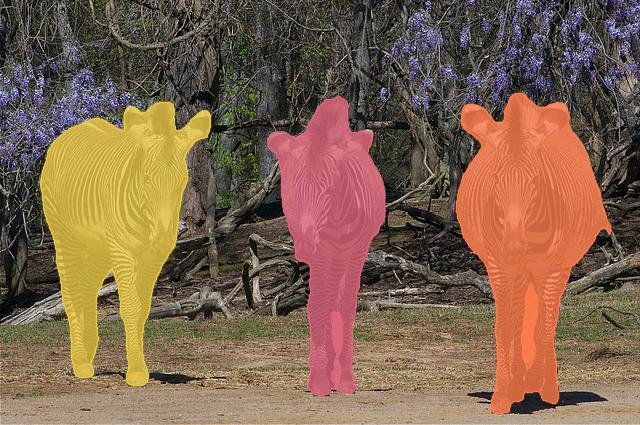} \\
    \end{tabular} \egroup
    \end{adjustbox}

    \begin{adjustbox}{width=\textwidth}
    \bgroup
    \def\arraystretch{0.2}
    \setlength\tabcolsep{0.2pt}
    \begin{tabular}{cccc}
    \includegraphics[height=2cm]{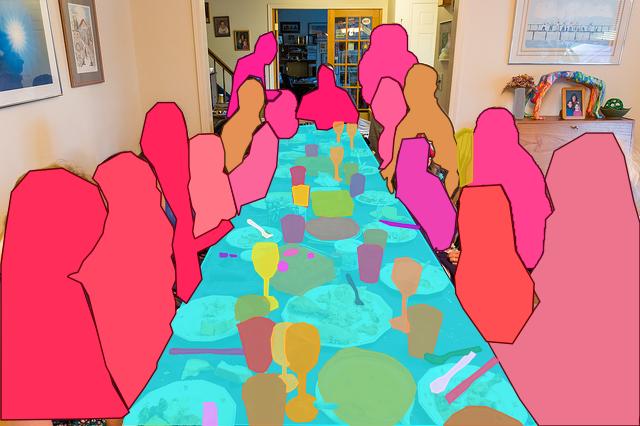} &
    \includegraphics[height=2cm]{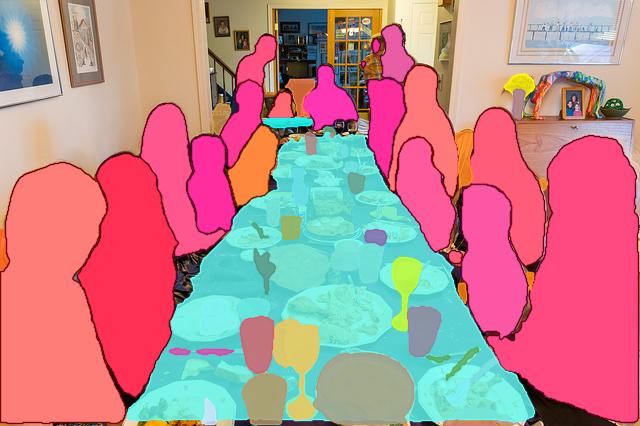} &
    \includegraphics[height=2cm]{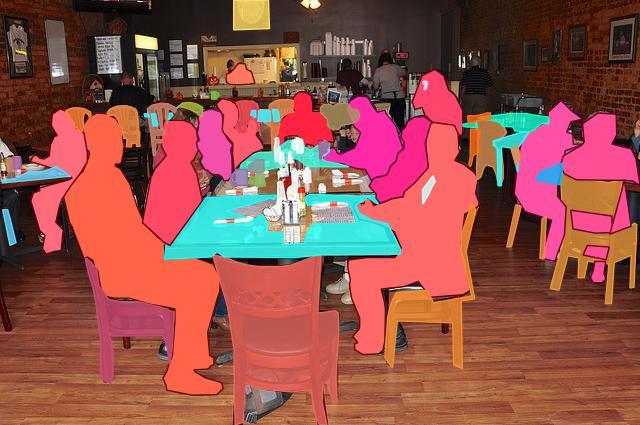} &
    \includegraphics[height=2cm]{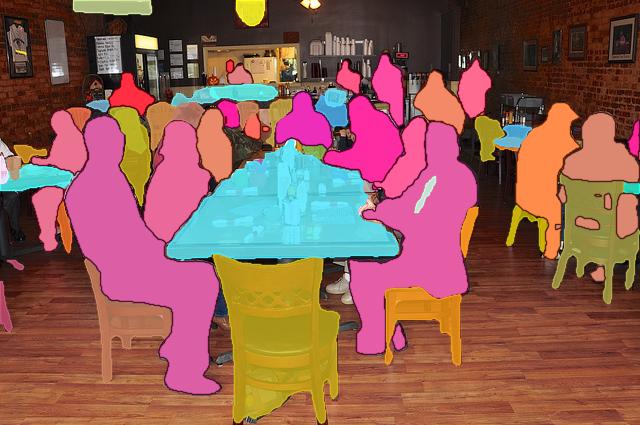} \\
    \end{tabular} \egroup
    \end{adjustbox}

    \begin{adjustbox}{width=\textwidth}
    \bgroup
    \def\arraystretch{0.2}
    \setlength\tabcolsep{0.2pt}
    \begin{tabular}{cccc}
    \includegraphics[height=2cm]{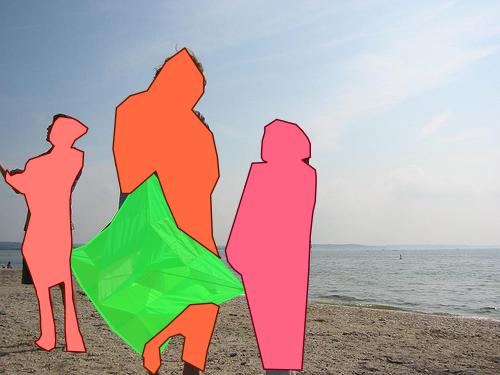} &
    \includegraphics[height=2cm]{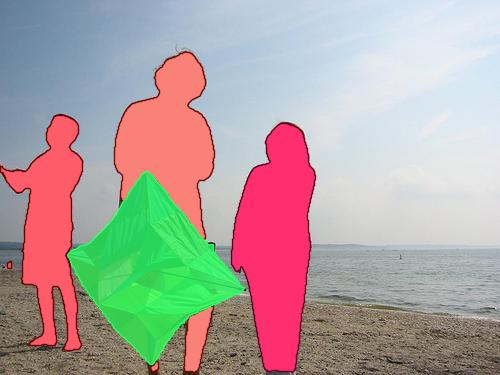} &
    \includegraphics[height=2cm]{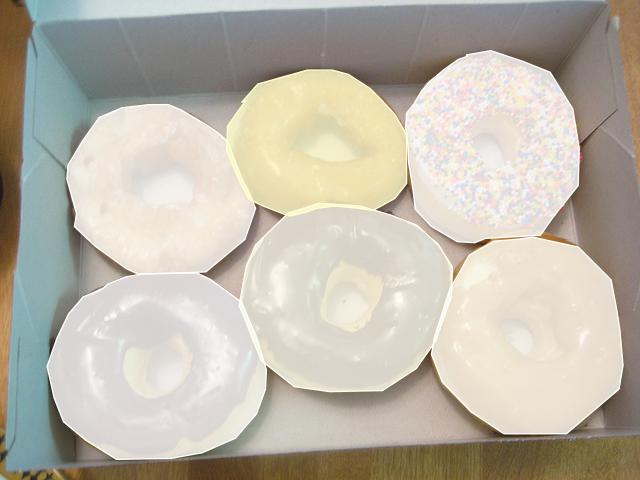} &
    \includegraphics[height=2cm]{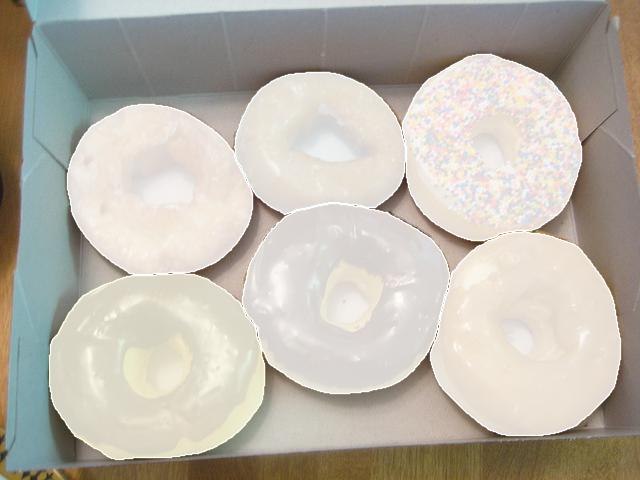} \\
    \end{tabular} \egroup
    \end{adjustbox}

    \begin{adjustbox}{width=\textwidth}
    \bgroup
    \def\arraystretch{0.2}
    \setlength\tabcolsep{0.2pt}
    \begin{tabular}{cccc}
    \includegraphics[height=2cm]{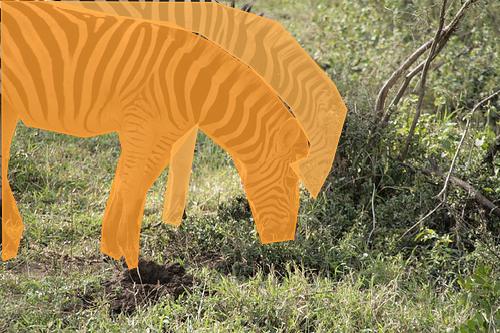} &
    \includegraphics[height=2cm]{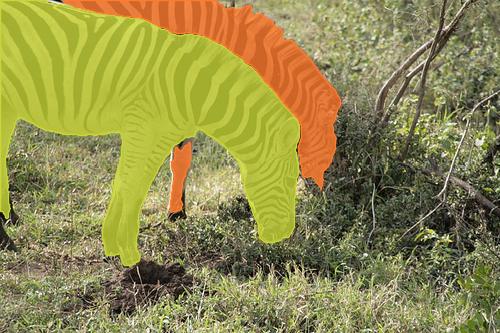} &
    \includegraphics[height=2cm]{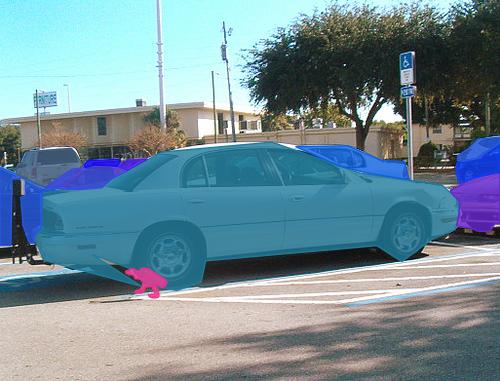} &
    \includegraphics[height=2cm]{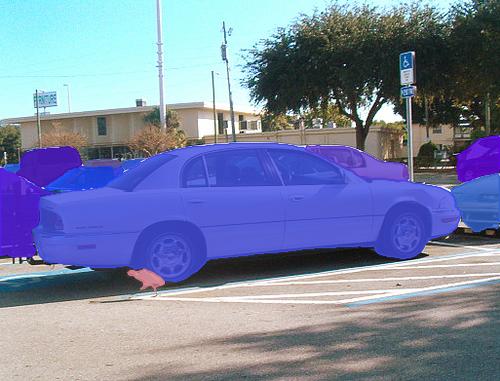} \\
    \end{tabular} \egroup
    \end{adjustbox}

    \begin{adjustbox}{width=\textwidth}
    \bgroup
    \def\arraystretch{0.2}
    \setlength\tabcolsep{0.2pt}
    \begin{tabular}{cccc}

    \includegraphics[height=2cm]{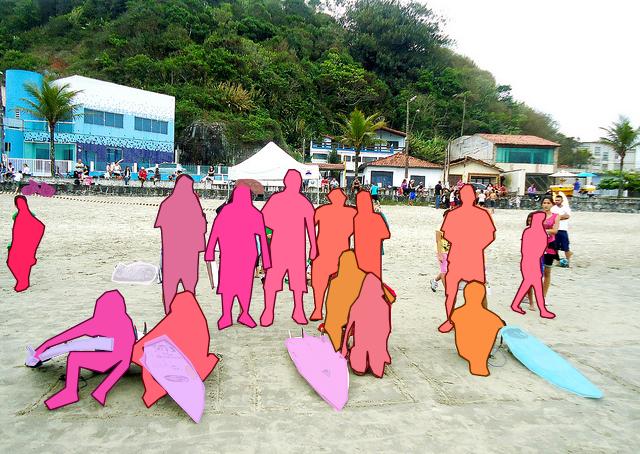} &
    \includegraphics[height=2cm]{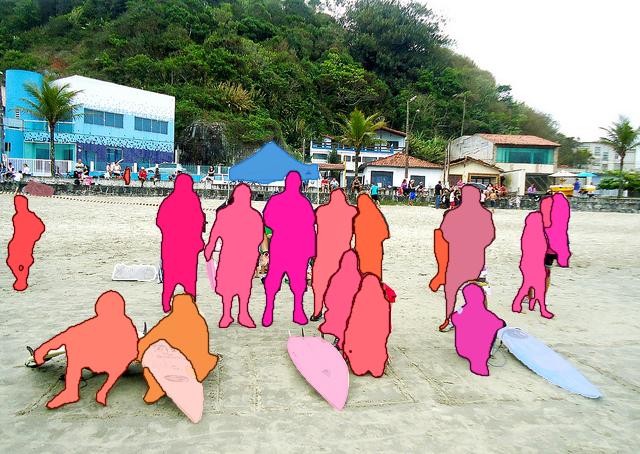} &
    \includegraphics[height=2cm]{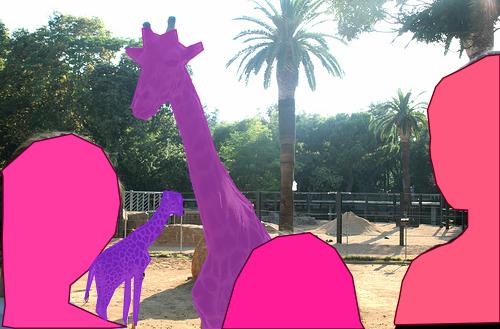} &
    \includegraphics[height=2cm]{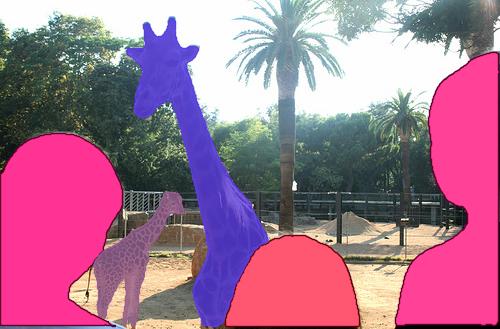} \\
    \end{tabular} \egroup
    \end{adjustbox}

    \begin{adjustbox}{width=\textwidth}
    \bgroup
    \def\arraystretch{0.2}
    \setlength\tabcolsep{0.2pt}
    \begin{tabular}{cccc}
    \includegraphics[height=2cm]{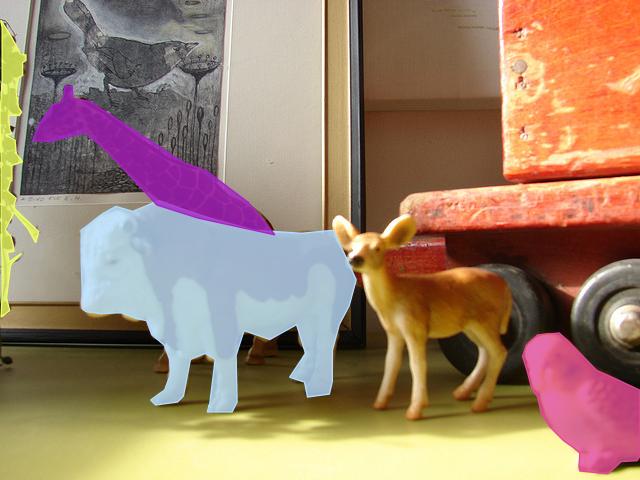} &
    \includegraphics[height=2cm]{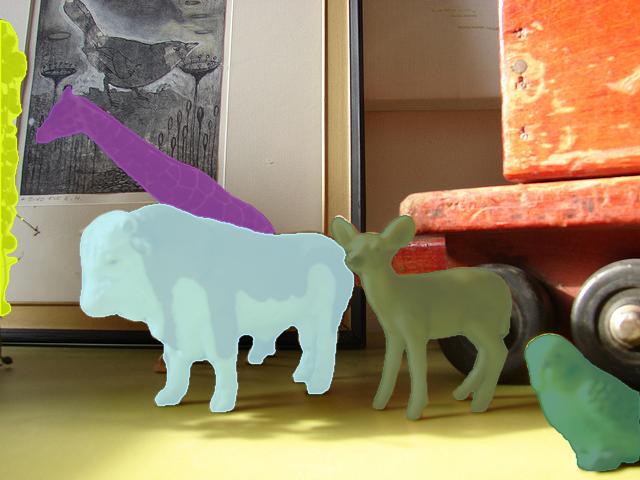} &
    \includegraphics[height=2cm]{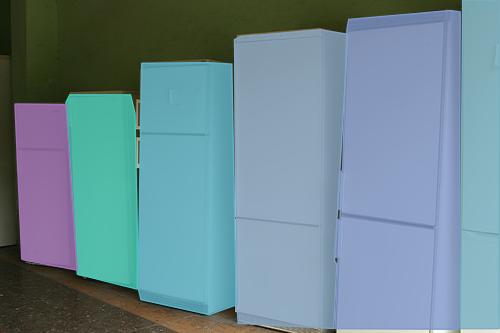} &
    \includegraphics[height=2cm]{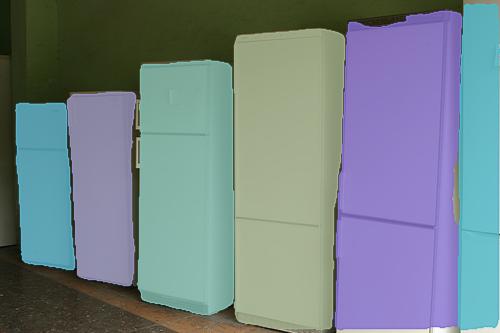} \\
    \end{tabular} \egroup
    \end{adjustbox}

  \caption{
  \textbf{Visualization of some predictions on the COCO dataset.} We use \modelname-D3 with a ResNet-50-d-DCN backbone that achieves 40.1 AP on the validation set with a speed of 32.5 FPS on a single V100 GPU. The first and third columns show the ground truth, and the second and fourth columns show the predictions. We set the confidence threshold to 0.5.
  }
  \label{fig:vis_insseg1}
\end{figure*}

\begin{figure*}[!t]
  \centering

  \begin{adjustbox}{width=\textwidth}
  \bgroup
  \def\arraystretch{0.2}
  \setlength\tabcolsep{0.2pt}
  \begin{tabular}{cccc}
  \includegraphics[height=2cm]{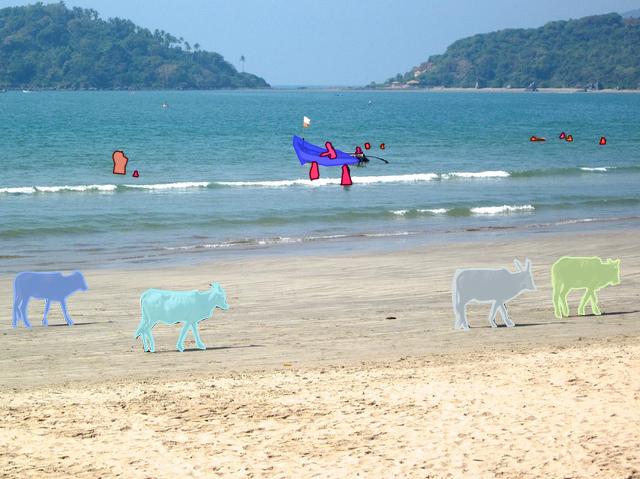} &
  \includegraphics[height=2cm]{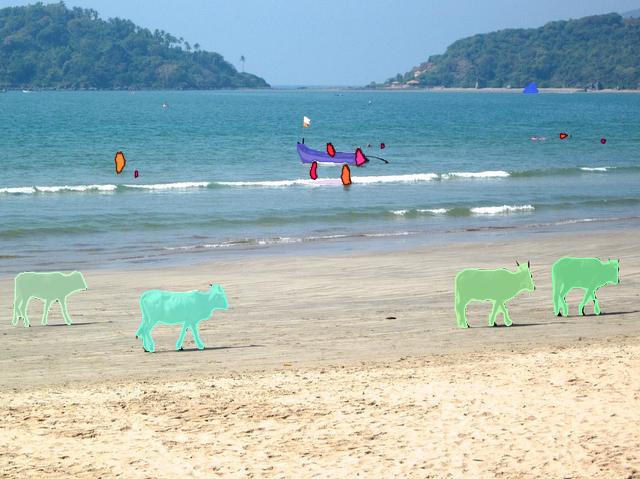} &
  \includegraphics[height=2cm]{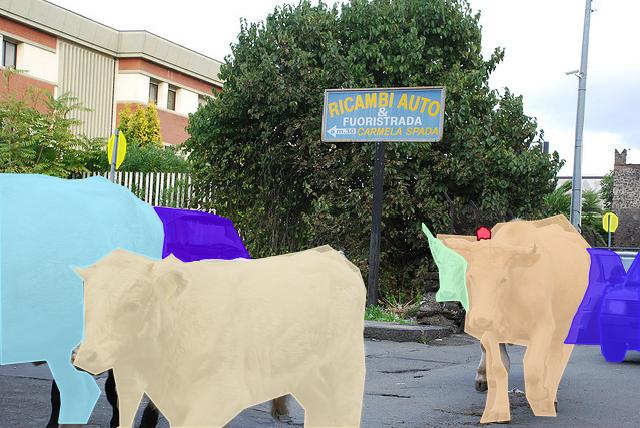} &
  \includegraphics[height=2cm]{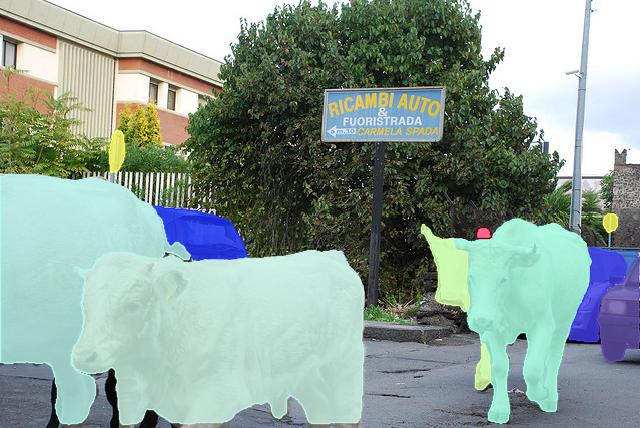} \\
  \end{tabular} \egroup
  \end{adjustbox}

  \begin{adjustbox}{width=\textwidth}
  \bgroup
  \def\arraystretch{0.2}
  \setlength\tabcolsep{0.2pt}
  \begin{tabular}{cccc}
  \includegraphics[height=2cm]{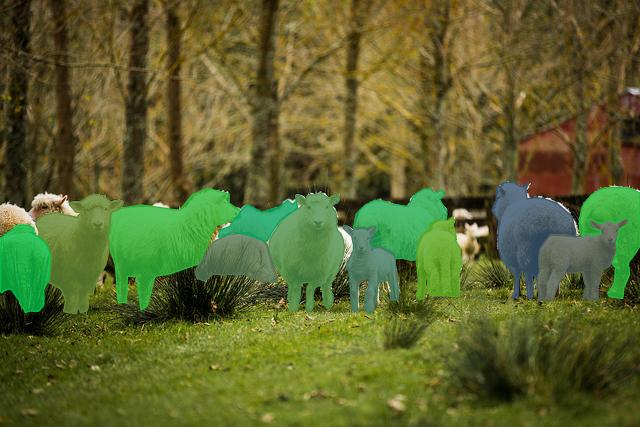} &
  \includegraphics[height=2cm]{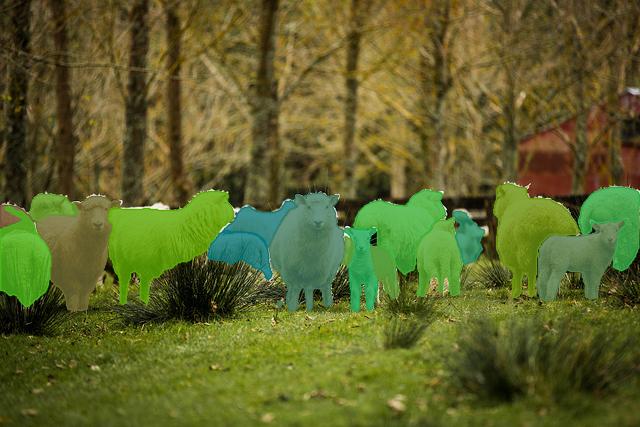} &
  \includegraphics[height=2cm]{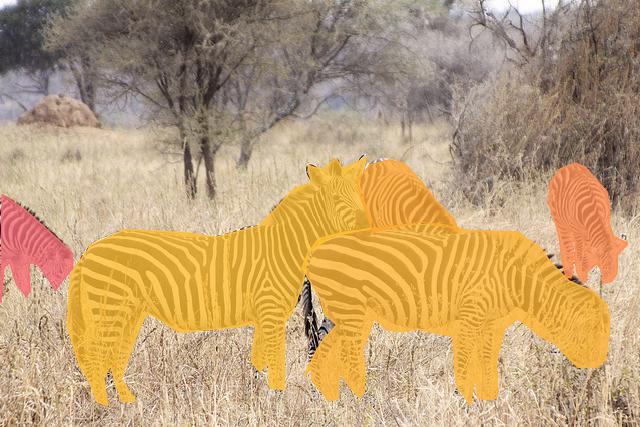} &
  \includegraphics[height=2cm]{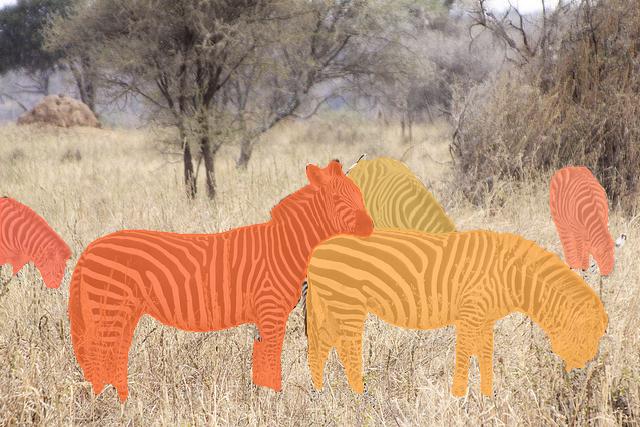} \\
  \end{tabular} \egroup
  \end{adjustbox}

  \begin{adjustbox}{width=\textwidth}
  \bgroup
  \def\arraystretch{0.2}
  \setlength\tabcolsep{0.2pt}
  \begin{tabular}{cccc}
  \includegraphics[height=2cm]{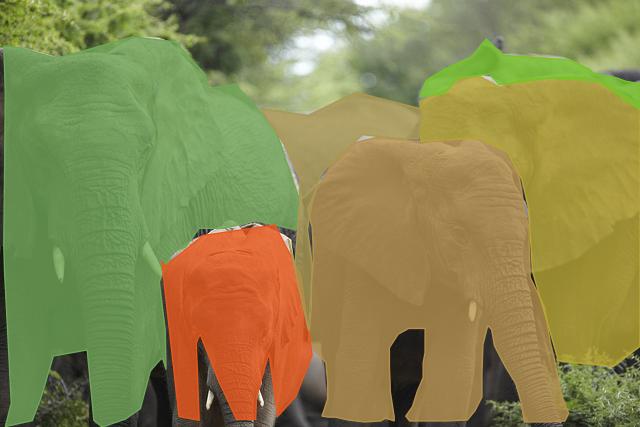} &
  \includegraphics[height=2cm]{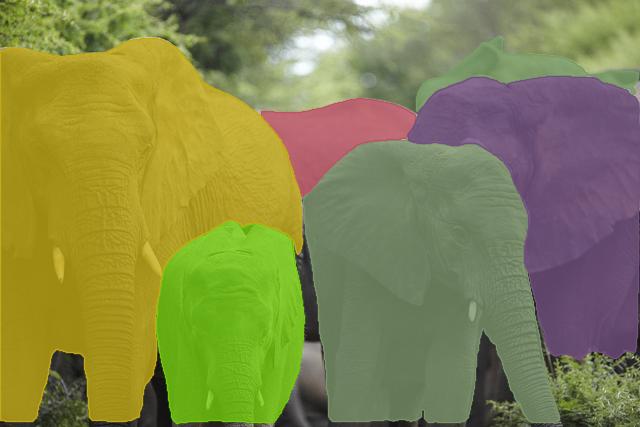} &
  \includegraphics[height=2cm]{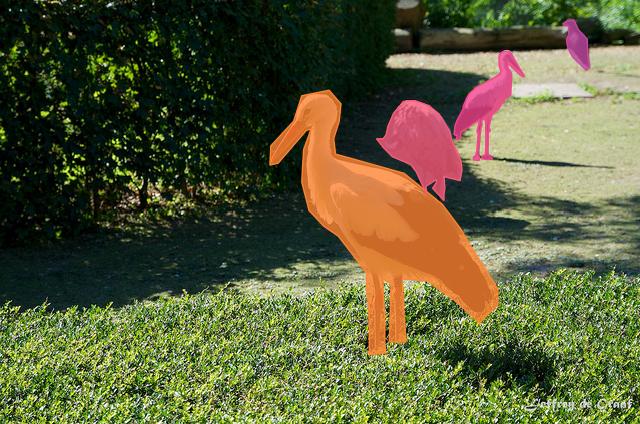} &
  \includegraphics[height=2cm]{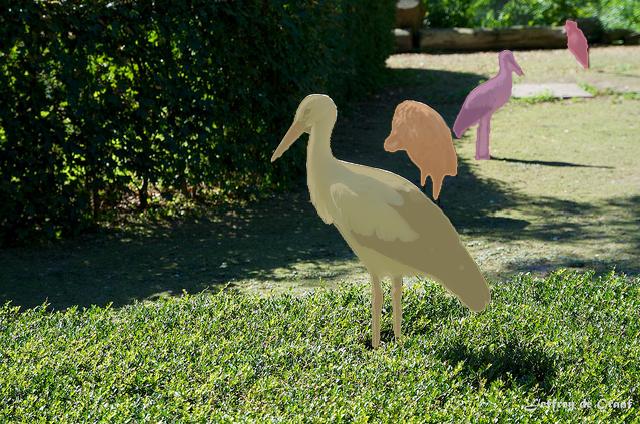} \\
  \end{tabular} \egroup
  \end{adjustbox}

  \begin{adjustbox}{width=\textwidth}
  \bgroup
  \def\arraystretch{0.2}
  \setlength\tabcolsep{0.2pt}
  \begin{tabular}{cccc}
  \includegraphics[height=2cm]{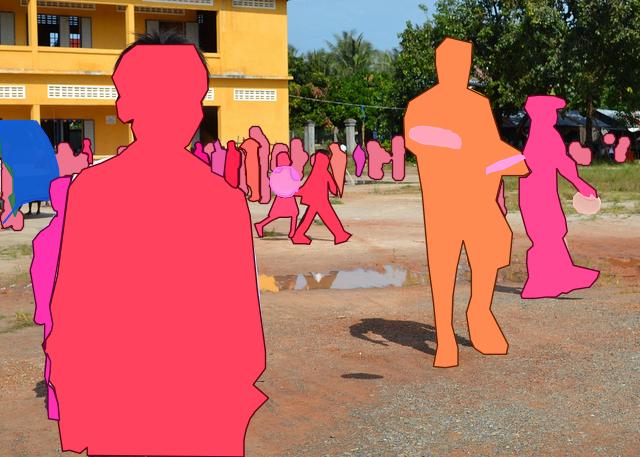} &
  \includegraphics[height=2cm]{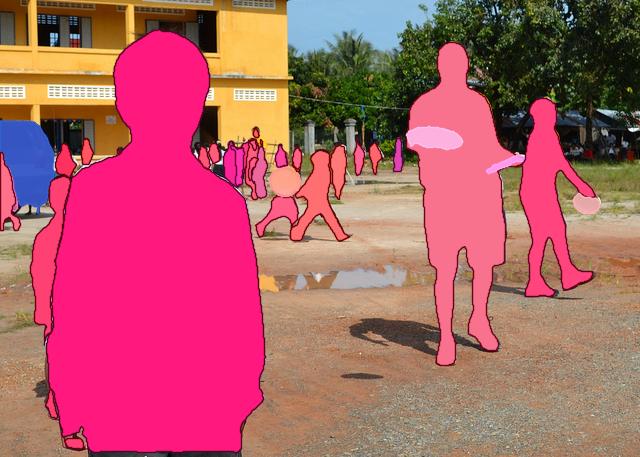} &
  \includegraphics[height=2cm]{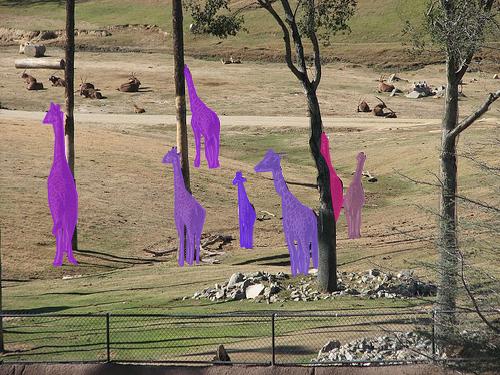} &
  \includegraphics[height=2cm]{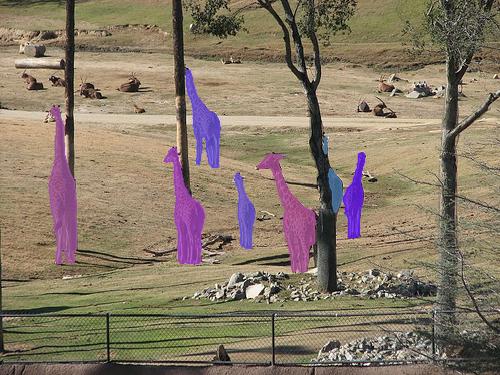} \\
  \end{tabular} \egroup
  \end{adjustbox}

  \begin{adjustbox}{width=\textwidth}
  \bgroup
  \def\arraystretch{0.2}
  \setlength\tabcolsep{0.2pt}
  \begin{tabular}{cccc}

  \includegraphics[height=2cm]{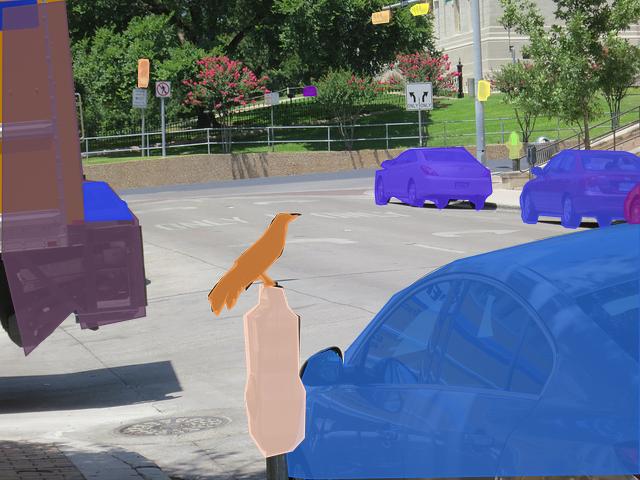} &
  \includegraphics[height=2cm]{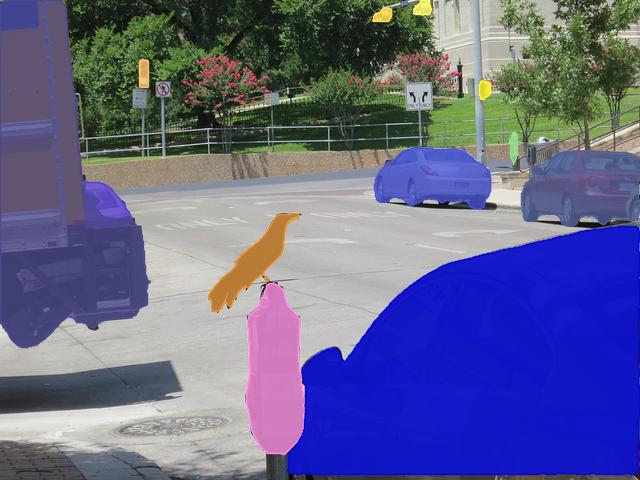} &
  \includegraphics[height=2cm]{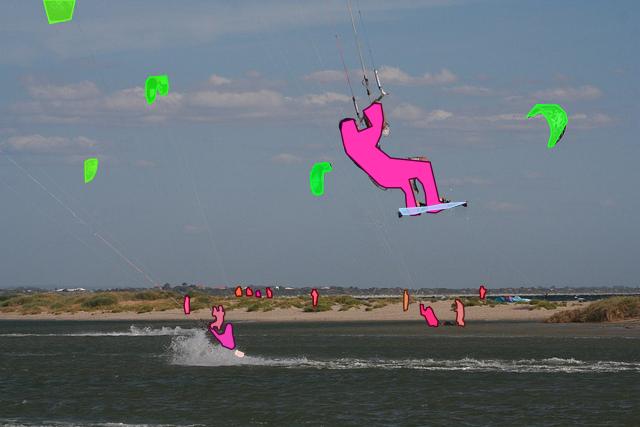} &
  \includegraphics[height=2cm]{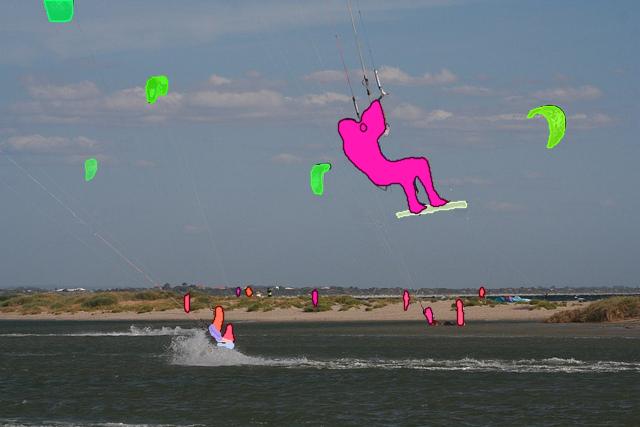} \\
  \end{tabular} \egroup
  \end{adjustbox}

  \begin{adjustbox}{width=\textwidth}
  \bgroup
  \def\arraystretch{0.2}
  \setlength\tabcolsep{0.2pt}
  \begin{tabular}{cccc}
  \includegraphics[height=2cm]{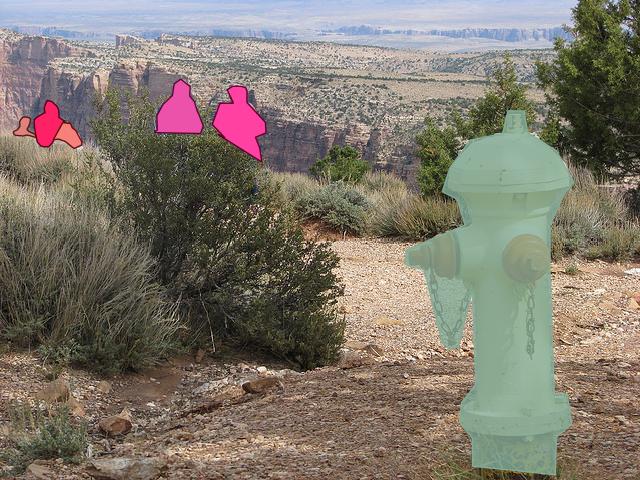} &
  \includegraphics[height=2cm]{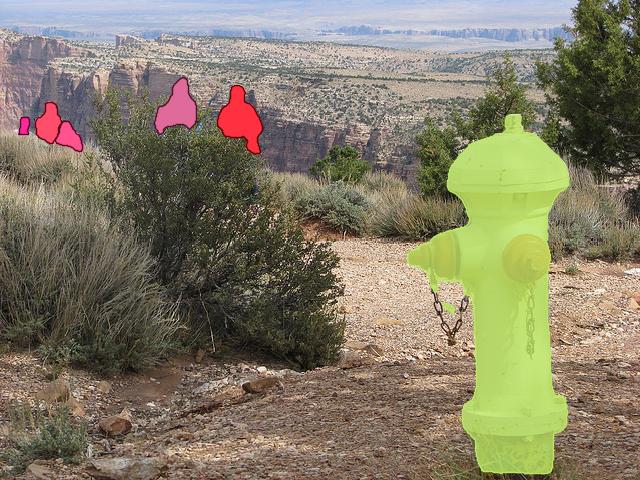} &
  \includegraphics[height=2cm]{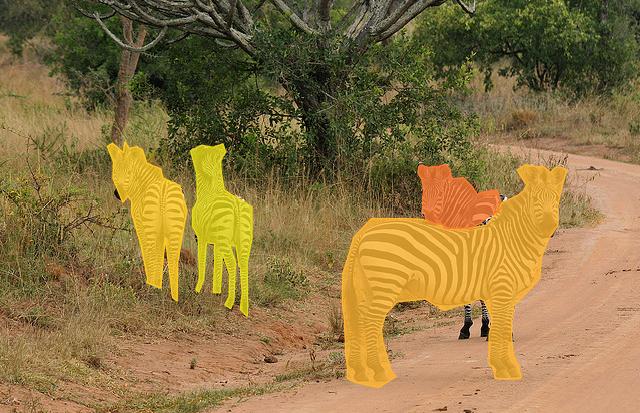} &
  \includegraphics[height=2cm]{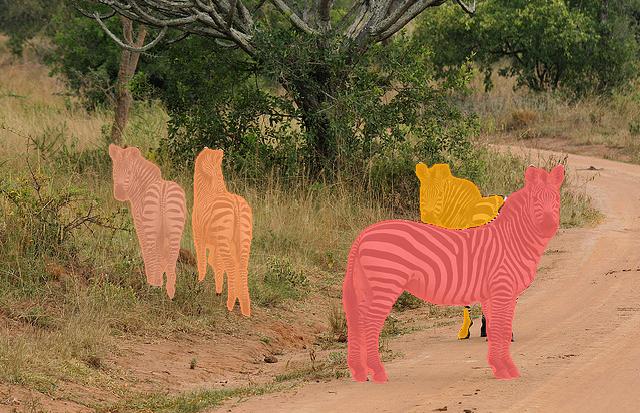} \\
  \end{tabular} \egroup
  \end{adjustbox}

\caption{
  \textbf{Visualization of another group of predictions on the COCO dataset.} We use \modelname-D3 with a ResNet-50-d-DCN backbone that achieves 40.1 AP on the validation set with a speed of 32.5 FPS on a single V100 GPU. The first and third columns show the ground truth, and the second and fourth columns show the predictions. We set the confidence threshold to 0.5.
}
\label{fig:vis_insseg2}
\end{figure*}

\begin{figure*}[!t]
  \centering

  \begin{adjustbox}{width=\textwidth}
  \bgroup
  \def\arraystretch{0.2}
  \setlength\tabcolsep{0.2pt}
  \begin{tabular}{cccc}
  \includegraphics[height=2cm]{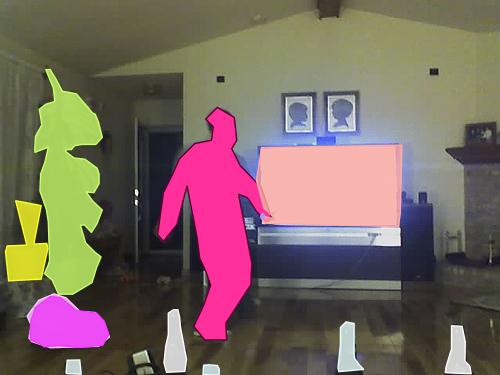} &
  \includegraphics[height=2cm]{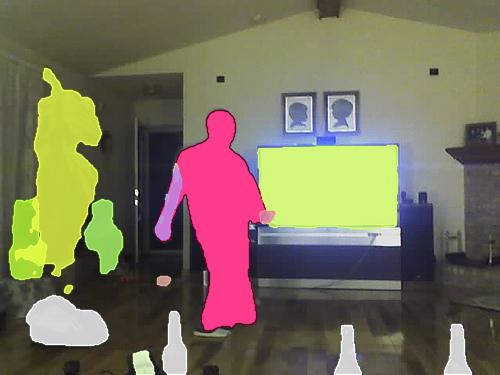} &
  \includegraphics[height=2cm]{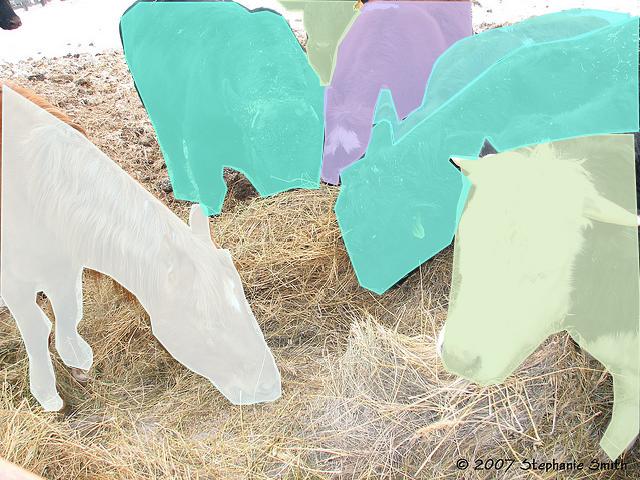} &
  \includegraphics[height=2cm]{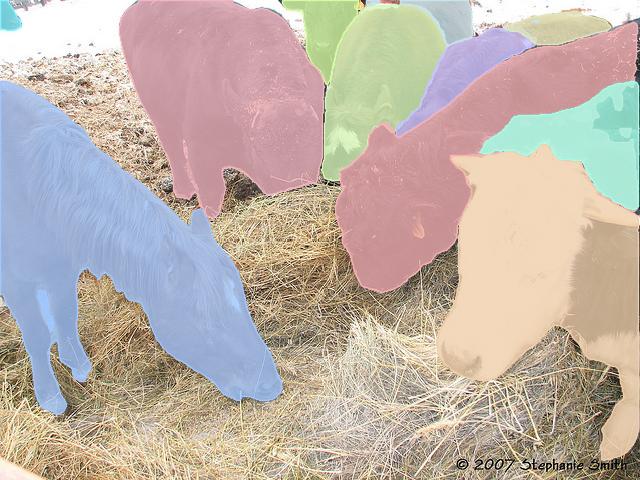} \\
  \end{tabular} \egroup
  \end{adjustbox}

\caption{
\textbf{Visualization of two typical failure cases on the COCO dataset.} Left: duplicate predictions (\eg, the person in the center). Right: over segmentation (\eg, the cow in the upper right corner). Also, there are a few false positive and false negative predictions (see the left sample result). Here the first and third columns are the ground truth, and the second and fourth columns are the failure predictions. The confidence threshhold is set to 0.5, as in \figref{fig:vis_insseg1} and \figref{fig:vis_insseg2}.
}
\label{fig:vis_insseg_failure_cases}
\end{figure*}

We visualize some predictions of the \modelname-D3 model with ResNet-50-d-DCN~\cite{he2016deep,he2019bag,zhu2019deformable} backbone on the COCO~\cite{lin2014coco} \texttt{val2017} set (40.1 AP) in \figref{fig:vis_insseg1} and \figref{fig:vis_insseg2}. \figref{fig:vis_insseg_failure_cases} shows two typical failure cases.

\end{document}